\documentclass[11pt]{article}

\usepackage[]{acl}

\usepackage[T1]{fontenc}
\usepackage[utf8]{inputenc}
\usepackage{times}
\usepackage{latexsym}
\usepackage{inconsolata}
\usepackage{microtype}

\usepackage{amsfonts,bbm}
\usepackage{amsmath,amssymb}

\usepackage{inconsolata}
\usepackage[normalem]{ulem}
\usepackage{cuted} 
\usepackage[dvipsnames]{xcolor} % master color package
\definecolor{nred}{RGB}{196, 38, 11}
\definecolor{ngreen}{RGB}{18, 141, 21}
\definecolor{nblue}{RGB}{41, 52, 190}
\definecolor{dartgreen}{HTML}{00693e}
\definecolor{maroon}{cmyk}{0,0.87,0.68,0.32}
\usepackage{graphicx}
\usepackage{subcaption}
\usepackage{wrapfig}
\usepackage{tikz}
\usetikzlibrary{shadows}

\usepackage{booktabs}
\usepackage{multirow}
\usepackage{tabularx}
\usepackage{makecell}
\usepackage{colortbl}
\usepackage{adjustbox}

\usepackage{algorithm}
\usepackage{algorithmicx}
\usepackage{algpseudocode}

\usepackage{enumitem}

\usepackage[most]{tcolorbox}
\tcbuselibrary{breakable, skins, theorems}
\usepackage{cleveref}

\usepackage{xspace}
\usepackage{pifont}
\usepackage{cprotect}
\usepackage{soul}
\usepackage{etoc}
\usepackage{minitoc}
\usepackage{titletoc}
\usepackage{caption}
\usepackage{nicefrac}

\title{\texttt{Cluster-R1}: Large Reasoning Models Are Instruction-following Clustering Agents}

\author{Peijun Qing\thanks{Work done during internship at Adobe Research}\textsuperscript{1}, Puneet Mathur\thanks{Primary Internship Mentor}\textsuperscript{2}, Nedim Lipka\textsuperscript{2}, Varun Manjunatha\textsuperscript{2}, Ryan Rossi\textsuperscript{2}, Franck Dernoncourt\textsuperscript{2}, \\ \bf{Saeed Hassanpour\textsuperscript{1}, Soroush Vosoughi\textsuperscript{1}}\\
\textsuperscript{1}\small{Dartmouth College, USA}\\
\textsuperscript{2}\small{Adobe Research, San Jose, USA}\\}

\definecolor{nred}{RGB}{196, 38, 11}
\definecolor{ngreen}{RGB}{18, 141, 21}
\definecolor{nblue}{RGB}{41, 52, 190}
\definecolor{dartgreen}{HTML}{00693e}
\definecolor{maroon}{cmyk}{0,0.87,0.68,0.32}
\newcommand*\circled[1]{%
  \tikz[baseline=(char.base)]{
    \node[
      shape=circle,
      draw,
      inner sep=0.2ex,
      line width=0.4pt,
      font=\normalsize
    ] (char) {#1};
  }%
}

\begin{document}
\maketitle

\begin{abstract}
General-purpose embedding models excel at recognizing semantic similarities but fail to capture the characteristics of texts specified
by user instructions.  In contrast, instruction-tuned embedders can align embeddings with textual instructions yet cannot autonomously infer latent corpus structures, such as determining the optimal number of clusters. To address both limitations, we reframe instruction-following clustering as a generative task and train large reasoning models (LRMs) as autonomous clustering agents. Our reasoning-driven training pipeline enables LRMs to interpret high-level clustering instructions and then infer the corresponding latent groupings. To evaluate this paradigm, we introduce \textsc{ReasonCluster}, a comprehensive benchmark comprising 28 diverse tasks spanning daily dialogue, legal cases, and financial reports. 
Experiments across diverse datasets and clustering scenarios show that our approach consistently outperforms strong embedding-based methods and LRM baselines, demonstrating that explicit reasoning fosters more faithful and interpretable instruction-based clustering.

%Across multiple datasets and clustering scenarios, our models consistently surpass strong embedding models and LRMs baselines

% General-purpose embedding models, such as Gemini Embeddings, are not designed to follow user-specific instructions, which limits their adaptability in instruction-guided representation tasks. In contrast, instruction-following embedding models, such as INBEDDER, can align embeddings with user-provided prompts but lack the ability to autonomously uncover latent structures in a new corpus—such as inferring the number of meaningful clusters. To address both limitations, we propose leveraging large reasoning models that can simultaneously interpret user instructions and dynamically analyze corpus structure. We introduce a benchmark comprising 40 tasks, significantly expanding beyond the 2 tasks included in existing benchmarks. Our approach integrates instruction-based embedding with high-level reasoning, enabling both guided representation learning and unsupervised structure discovery within a unified framework. 

\end{abstract}

\section{Introduction}

Text clustering is a cornerstone of natural language processing (NLP) and data analysis pipelines, supporting diverse applications such as identifying public sentiment in social media~\citep{PARK2022103524}, diagnosing accident causes~\cite{xu2022cause}, assisting data synthesis~\cite{zeng2025evaltree}, and improving tool-use efficiency for tool-enhanced large language model (LLM)~\citep{liu2025toolplanner}. By grouping related items into coherent structures, clustering enables the scalable organization and interpretation of unstructured information, thereby improving the efficiency and effectiveness of downstream tasks.

% Clustering often constitutes the first step in data analysis and NLP pipeline construction,  By grouping related items into coherent structures, clustering enables scalable organization and interpretation of unstructured information, thereby enhancing both the efficiency and effectiveness of downstream tasks.

\begin{figure}[t]
    \centering
    \includegraphics[scale=0.65]{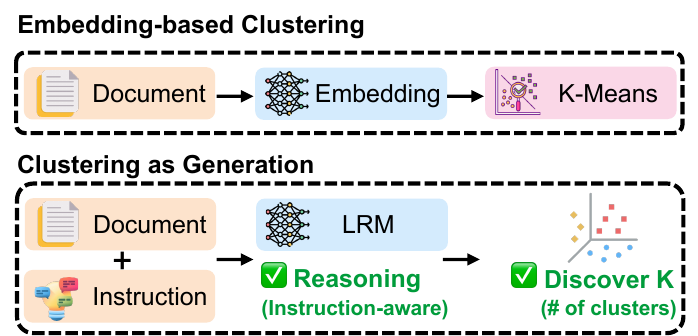}
    \caption{Overview of embedding-based vs. reasoning-driven clustering. LRMs follow diverse user instructions and adaptively infer latent group structures.}
    \label{fig:intro}
    \vspace{-4mm}
\end{figure}

A common paradigm for text clustering applies unsupervised algorithms such as K-means~\citep{macqueen1967kmeans} or Gaussian Mixture Models (GMMs)~\citep{dempster1977gaussian} on top of embedding vectors produced by pre-trained encoders (e.g., SentenceBert~\citep{reimers-gurevych-2019-sentencebert}), as shown in Figure~\ref{fig:intro}. These embedding models, trained primarily with contrastive objectives, excel at capturing general semantic similarity and yield strong performance in standard text embedding benchmarks~\citep{muennighoff2022mteb}. However, such representations often fail when clustering tasks demand alignment with user-specific objectives that transcend generic semantics~\citep{instructor, answer-need}. 

Recent instruction-following embedding models address this gap by integrating natural-language instructions into the embedding process, enabling embeddings to align more closely with user intent~\citep{instructor, answer-need, qwen3-embedding}. Models such as Instructor~\citep{instructor} and InBedder~\citep{answer-need} fine-tune text encoders with instruction-conditioned contrastive objectives, yielding improved adaptability to downstream goals. However, these methods remain fundamentally limited: they depend on external clustering algorithms to uncover latent corpus structure, leaving the model incapable of autonomously determining the number of clusters and their labels.

In this study, we address the above limitations by \textbf{reframing instruction-following clustering as a generation task executed by a large reasoning model}. As shown in Figure \ref{fig:intro}, given a set of indexed texts and an instruction describing the clustering goal, the LRM first infers the number of clusters consistent with the instruction, and then generates the cluster assignments. This reframing allows the model to internalize both the reasoning process and the clustering output within a single generation process. To enable this capability, we propose a post-training framework that integrates reasoning distillation and reinforcement learning (RL). First, we perform supervised fine-tuning (SFT) on synthesized reasoning traces that teach the model to verbalize its clustering logic and produce coherent, structured outputs. Then we apply Group Relative Policy Optimization (GRPO)~\citep{shao2024grpo} with hybrid reward that jointly enforces adherence to output format, accurate cluster-count prediction, and high clustering quality.\looseness-1 

% Given an indexed corpus and a clustering instruction, the model first reasons about potential cluster structures and then \emph{generates} both (i) the inferred number of clusters consistent with the instruction and (ii) the assignment of each text to the corresponding cluster. This paradigm unifies two capabilities that have traditionally been separated: (a) instruction comprehension and user-intent alignment, characteristic of instruction-following embedders~\citep{answer-need, instructor}, and (b) autonomous discovery of clustering taxonomies, typically absent from embedding-only pipelines.

% This dual-stage training significantly enhances reasoning consistency, reduces hallucinations, and strengthens the link between generated reasoning and measurable clustering performance.

% To systematically evaluate this paradigm, we construct a comprehensive instruction-following clustering benchmark that spans multiple domains (e.g., legal, finance) and diverse clustering intents (e.g., topical, functional, demographic)~\citep{chalkidis-etal-2019-echr, zheng2023lmsys}.
To systematically evaluate this paradigm, we construct a comprehensive instruction-following clustering benchmark spanning multiple domains and diverse clustering intents. 
We compare our models against a wide array of baselines, including both proprietary and open-source LLMs/LRMs, as well as conventional and instruction-tuned embedding models. Across all settings, our approach consistently achieves the highest performance, demonstrating that explicit reasoning, when trained with task-aligned rewards, yields superior, instruction-faithful clustering. In summary, our main contributions are as follows:
\begin{itemize}[leftmargin=*]
    \item We re-conceptualize \textbf{clustering as a generation task}, where the model jointly infers cluster structure and assigns items accordingly, addressing the challenge of instruction-following and reasoning-intensive clustering.

    \item We present \textsc{\textbf{REASONCLUSTER}}, a comprehensive benchmark for systematic evaluation of instruction-following clustering spanning multiple domains and clustering intents.

    \item We propose \texttt{\textbf{Cluster-R1}}, a post-training recipe that combines distillation with GRPO to optimize instruction-based clustering that outperforms powerful reasoning models like GPT-o3 by 3-5\%  across benchmarks.
\end{itemize}

% \paragraph{Contributions.}
% \begin{itemize}
% \item \textbf{New paradigm.} We formulate \emph{instruction-following clustering as generation}: an LRM jointly infers the number of clusters and produces index-level assignments conditioned on user goals.
% \item \textbf{Training recipe.} We present a two-stage pipeline—reasoning-trace SFT followed by GRPO with hybrid rewards—that improves format fidelity, cluster-count accuracy, and taxonomy quality.
% \item \textbf{Comprehensive evaluation.} We benchmark against cutting-edge LLM/LRM systems and both general-purpose and instruction-following embedders, showing consistent gains in V-measure and robustness across data regimes.
% \end{itemize}

\begin{figure}[ht]
  \centering
  \begin{tcolorbox}[
      colback=gray!5!white,
      colframe=dartgreen,
      title=Clustering as Generation Task,
      boxrule=0.3mm,
      arc=3mm,
      width=\columnwidth
    ]
\small
\textbf{System Prompt:}

You are a clustering assistant to do text clustering. Given a clustering goal and a list of indexed corpus:
First, read through all texts and think how can they be clustered based on the goal,
determine the total number of clusters. Then think about how to assign all texts into these clusters. Check the answer format before giving the final answer: every item must be assigned to
exactly one cluster, and no item should appear in multiple clusters or be missing. The reasoning and answer must be enclosed within \texttt{<think>} \texttt{</think>} and \texttt{<answer>} \texttt{</answer>} tags, respectively. \\

Final Output Format should be:

\texttt{<think>} assistant's reasoning process here \texttt{</think>}

\texttt{<answer>}

Total clusters: [N].\\
cluster1: [item\_numbers separated by commas].\\
cluster2: [item\_numbers separated by commas].\\
...

\texttt{</answer>}\\

Now, please follow the format for the following clustering task:

Goal: \textcolor{dartgreen}{\texttt{\{clustering instruction\}}}

Text: \textcolor{dartgreen}{\texttt{\{enumerated texts\}}}

% Example:

% Clustering Goal: \{Please do clustering based on the intent\}

% Text: \\
% 1. My card has not arrived yet… \\ 
% 2. I ordered a card a week ago, and it's still not here. What do I do? \\  
% ...  
% \textbf{User:}  
% \{joined\_text\}  
% \textbf{Assistant:}  
% \texttt{<think>}

% \textbf{Output:}  
% \begin{verbatim}
% <think> reasoning process here </think>
% \end{verbatim}
% \begin{verbatim}
% <answer> 
% Total clusters: [2].
% cluster1: [1, 3, 5]. cluster2: [2, 4, 6]. 
% </answer>
% \end{verbatim}

  \end{tcolorbox}
  \vspace{-5pt}
  \caption{The system prompt for clustering.}
  \vspace{-15pt}
  \label{fig:prompt_clustering}
  %\vspace{-10pt}
\end{figure}

% \begin{table}[!]
% \centering  
% \resizebox{\textwidth}{!}{
% \tiny
% \begin{tabular}{lccccc}
% \toprule
% % Metric & \makecell{StructRAG\\vs.\\Long-context} & \makecell{StructRAG\\vs.\\RAG} & \makecell{StructRAG\\vs.\\RQ-RAG} & \makecell{StructRAG\\vs.\\GraphRAG} \\
% % \midrule
% % Comprehensiveness& & &  & \\ 
% % Diversity & & & & \\ 
% % Empowerment & & & & \\ 
% % Directness & & & & \\ 
% \textbf{Method} & \textbf{Instruction-following} & \textbf{Discover K} & \textbf{CLustering }  \\
% \midrule
% Gemini& No & No &  \\ 

% \bottomrule
% \end{tabular}
% }

% \label{table:win}
% \end{table}
\begin{figure*}[!ht]
    \centering
    \includegraphics[width=2.1\columnwidth]{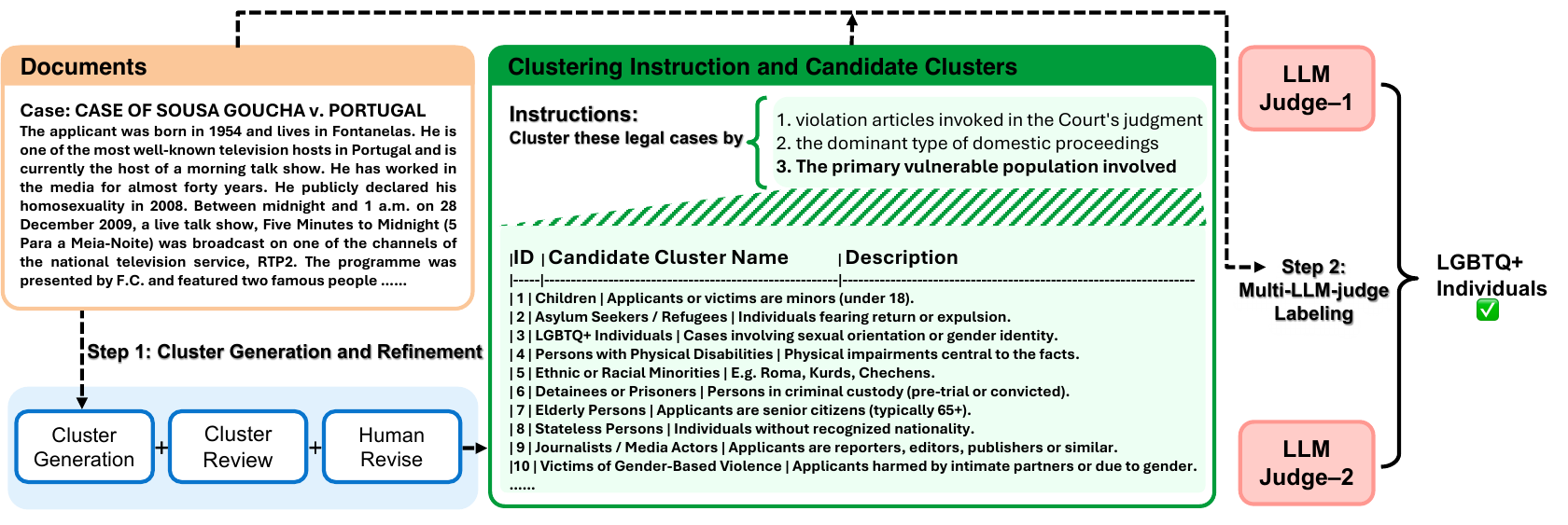}
    \caption{ The overview of our multi-agent data synthesis pipeline.
    }
    \label{fig: data_syn_multiagent}
    %\vspace{-15pt}
\end{figure*}

% \section{Related Work}
% \label{sec: related}

% Classical text clustering methods rely on fixed representations combined with unsupervised algorithms such as K-means, Gaussian mixture models, or hierarchical clustering~\citep{macqueen1967some, dempster1977maximum, johnson1967hierarchical}. Although widely used, these approaches assume a static clustering objective and cannot adapt to user-specified instructions. Instruction-following embedders address this limitation by conditioning representations on natural language prompts~\citep{instructor, answer-need}, but still treat clustering as a downstream algorithmic step, requiring externally specified cluster counts and offering limited capacity for structural reasoning.

% More recent work explores large language models as active agents in clustering, leveraging natural language explanations, constraints, or feedback to guide grouping decisions~\citep{zhang-etal-2023-clusterllm, wang2023goal, viswanathan2024large}. Our work builds on this line of research by reframing instruction-following clustering as a fully generative reasoning task, in which the model autonomously infers both the number of clusters and the cluster assignments. A detailed discussion of related work is provided in Appendix~\ref{appen: related}.

\section{Related Work}
\label{sec: related}

\noindent\textbf{Instruction-following embedders.} General-purpose embedding models primarily capture task-agnostic semantic similarity~\citep{mikolov2013efficient,pennington2014glove,devlin2018bert,reimers-gurevych-2019-sentence,gao-etal-2021-simcse,all_minilm_l6_v2_card}, making them ill-suited for clustering under user-specified criteria. Instruction-tuned embedders~\citep{instructor,answer-need,qwen3-embedding} incorporate natural-language prompts to better align representations with user intent, but still depend on external clustering algorithms with manually chosen hyperparameters and limited capacity to infer cluster structure. We instead cast clustering as an end-to-end generative task, training LRMs to infer both the cluster structure and grouping logic given the instruction.

% General-purpose embedding models learn task-agnostic similarity~\citep{mikolov2013efficient,pennington2014glove,devlin2018bert,reimers-gurevych-2019-sentence,gao-etal-2021-simcse,all_minilm_l6_v2_card}, making it difficult to align representations with user-specified criteria. Instruction-following embedders condition embeddings on natural-language instructions~\citep{instructor,answer-need,qwen3-embedding}, but still rely on heuristic clustering algorithms with manually specified hyperparameters and limited structural reasoning. We instead train an LRM to infer the cluster grouping implied by the instruction through an end-to-end generation fashion. %including both $K$ and assignments.
%In contrast, we train LRMs to directly infer instruction-implied cluster grouping in an end-to-end generative framework.

% \noindent\textbf{LLM-involved clustering.}
% More recent work explores large language models as active agents in clustering, leveraging natural language explanations, constraints, or feedback to guide grouping decisions~\citep{zhang-etal-2023-clusterllm, wang2023goal, viswanathan2024large}. Our work builds on this line of research by reframing instruction-following clustering as a fully generative reasoning task, in which the model autonomously infers both the number of clusters and the cluster assignments. A detailed discussion of related work is provided in Appendix~\ref{appen: related}.

\noindent\textbf{LLM-involved clustering.}
Prior works have explored the use of LLMs for prompt-based clustering, leveraging natural-language descriptions, constraints, or iterative refinement to guide cluster grouping decisions~\citep{zhang-etal-2023-clusterllm,wang2023goal,viswanathan2024large,feng2024llmedgerefine,de2023idas,tipirneni2024context,nakshatri2023using}. However, they fail to utilize their reasoning capabilities to jointly infer the clusters and the assignments. Ours is the first work that formulates instruction-following clustering as an end-to-end generative task performed by a single LLM. See Appendix~\ref{appen: related} for a broader discussion.

% Prior work on text clustering spans classical unsupervised methods, instruction-conditioned representation learning, and LLM-assisted clustering. Classical approaches combine fixed representations with algorithms such as K-means or Gaussian mixture models, optimizing geometric or probabilistic objectives independent of user intent~\citep{macqueen1967some, dempster1977maximum}. 
% Instruction-following embedders condition representations on natural language prompts to improve downstream alignment~\citep{instructor, answer-need}, but still decouple representation learning from cluster structure inference. More recent methods leverage large language models to provide constraints, explanations, or refinements for clustering~\citep{zhang-etal-2023-clusterllm, viswanathan2024large}. Our work fits within this emerging direction by treating instruction-following clustering as a generative reasoning process. A detailed discussion is provided in Appendix~\ref{appen: related}.

\section{Problem Definition}
\label{subsec: tasks definition}
As illustrated in Figure~\ref{fig:prompt_clustering}, we treat instruction-following clustering as a generative task. The input consists of a single clustering instruction along with a list of texts, and the output involves the cluster number and the assignment of clusters.

% Unlike conventional unsupervised clustering, which depends solely on intrinsic data similarities, this formulation integrates semantic guidance from user-provided instructions, thereby enabling adaptive and interpretable organization of data points.

Formally, let \(\mathcal{D} = \{ \mathbf{x}_1, \mathbf{x}_2, \dots, \mathbf{x}_n \}\) denote a dataset of \(n\) texts, where each \(\mathbf{x}_i \in \mathcal{X}\) is a token sequence from the input space \(\mathcal{X}\). Let \(I \in \mathcal{I}\) represent a natural language instruction drawn from the instruction space \(\mathcal{I}\), describing the intended clustering objective. The task is to produce a clustering \(\mathcal{C} = \{ C_1, C_2, \dots, C_k \}\) that partitions \(\mathcal{D}\) into \(k\) non-overlapping subsets. In contrast to traditional methods, the number of clusters \(k\) is not pre-specified but instead inferred from the instruction and corpus, i.e., \(k = f(I, \mathcal{D})\), where \(f\) denotes a reasoning process that aligns latent corpus structure with user intent.\looseness-1

\section{Benchmark Construction}
\label{sec: bench construction}

In this section, we present \textsc{ReasonCluster}, a benchmark designed to evaluate whether clustering methods can group documents according to diverse user instructions. 
%that demand reasoning extends beyond mere lexical or semantic similarity. 
Sec~\ref{subsec: data syn multi-agent} describes the multi-agent data synthesis pipeline, Sec~\ref{subsec: datasets} explains the data collection across three domains, Sec~\ref{subsec: c0c1c2 construction} presents the details of constructing the training set and evaluation set.

\subsection{Multi-Agent Data Synthesis}
\label{subsec: data syn multi-agent}
% We design a weakly supervised multi-agent pipeline with human-in-the-loop that balances scalability with reliability. As shown in Figure~\ref{fig: data_syn_multiagent}, the synthesis process includes three stages: instruction and cluster generation, cluster refinement, and consensus-based labeling.

Our data synthesis pipeline consists of three stages: cluster generation, cluster refinement, and consensus-based multi-agent labeling. Figure~\ref{fig: data_syn_multiagent} visualizes this process using a concrete legal case example, which we reference below to clarify the role of each agent.

\paragraph{Cluster Generation.}
Given a dataset comprising heterogeneous corpora, we first create candidate clustering instructions that emphasize reasoning-intensive perspectives and their associated clusters. As shown in Figure~\ref{fig: data_syn_multiagent} (step 1), given a set of legal case documents, the generation agent proposes candidate clustering instructions (e.g., “cluster legal cases by primary vulnerable population involved”) together with an initial set of candidate clusters, such as “LGBTQ+ Individuals” or “Children.” The system prompt of the generation agent is provided in Figure~\ref{fig:prompt_inst_taxonomy}. This stage yields an initial hypothesis space of possible instructions and clusters, which is then rewritten and refined by human annotators.

% For each dataset, we present a concise description together with a small set of randomly sampled, representative text examples to a generation agent. The agent is tasked with two objectives: (i) to propose clustering instructions that emphasize distinct reasoning-based perspectives, and (ii) to generate corresponding candidate cluster taxonomies aligned with those instructions. 

\paragraph{Cluster Refinement.}
% The structured outputs of the generation agent are then passed to a refinement agent, which enforces formal desiderata of candidate clusters. Specifically, the refine agent evaluates and revises candidate clusters to guarantee (i) \emph{mutual exclusivity}: categories must not overlap or contradict each other, and (ii) \emph{collective exhaustiveness}: categories together should cover the full space of possible intents in the given domain. The refinement process then involves human re-writing to ensure uniform granularity and eliminate vague or catch-all categories. This two-layer design reduces spurious and boundary clusters and generate a refined cluster set with name and description (green box in the ~\ref{fig:prompt_inst_taxonomy}). The system prompt for this stage is shown in Figure~\ref{fig:prompt_taxonomy_refine}.

The generation agent’s outputs are passed to a refinement agent that enforces two formal constraints: (i) \emph{mutual exclusivity}, ensuring non-overlapping categories, and (ii) \emph{collective exhaustiveness}, ensuring full coverage of relevant intents within the domain. Light human rewriting further standardizes granularity and removes vague categories. This agent yields a clean cluster taxonomy with explicit names and descriptions (green box in Figure~\ref{fig:prompt_inst_taxonomy}). The refinement prompt is shown in Figure~\ref{fig:prompt_taxonomy_refine}.

% A secondary review agent evaluates consistency and coverage, while human annotators selectively intervene in cases of ambiguity or disagreement.

\paragraph{Consensus-Based Multi-Agent Labeling.}
After a refined cluster set is finalized, two independent labeling agents assign each document (e.g., the Sousa Goucha case) to one cluster based on the refined set, and only consensus assignments are retained, as shown in Figure~\ref{fig:prompt_inst_taxonomy} (step 2). The conflicting cases are discarded. This mechanism provides two advantages: it significantly reduces manual labeling overhead while avoiding ambiguous or boundary cases. 
The system prompt for this stage is shown in Figure~\ref{fig:prompt_classification}.
% After a provisional cluster set is finalized, we adopt a consensus-driven multi-agent labeling framework based on the LLM-as-a-judge paradigm. Each text instance is classified according to the refined cluster set using a standardized classification prompt (Figure~\ref{fig:prompt_classification}). Two independent labeling agents assign cluster labels, and a consensus protocol determines acceptance: samples with consistent labels across agents are retained automatically, while 

% The result is a dataset that combines scalability, interpretability, and high reliability, achieved through the complementary strengths of multi-agent collaboration and selective human validation.

\begin{figure*}[t!]
\centering

\begin{subfigure}[t]{0.32\linewidth}
    \centering
    \raisebox{60pt}{
    \resizebox{\linewidth}{!}{
        \begin{tabular}{llccc}
        \toprule
        \textbf{Source} & \textbf{Split} & \textbf{text len} & \textbf{input len} & \textbf{\#data} \\
        \midrule
        \multirow{3}{*}{ECHR}
          & C0 & 696.4 & 9872.4 & 150 \\
          & C1 & 687.9 & 10239.9 & 150 \\
          & C2 & 959.6 & 12643.6 & 132 \\
        \midrule
        \multirow{3}{*}{LMSY}
          & C0 & 410.5 & 9237.8 & 500 \\
          & C1 & 411.3 & 9042.7 & 500 \\
          & C2 & 478.1 & 7201.9 & 250 \\
        \midrule
        \multirow{3}{*}{SP500}
          & C0 & 772.5 & 13420.6 & 179 \\
          & C1 & 809.1 & 13810.5 & 170 \\
          & C2 & 836.6 & 14405.6 & 140 \\
        \bottomrule
        \end{tabular}
    }}
    
    \caption{Evaluation dataset statistics.}
    \label{fig:subfig-table}
\end{subfigure}
\hfill
\begin{subfigure}[t]{0.32\linewidth}
    \centering
    \includegraphics[width=\linewidth]{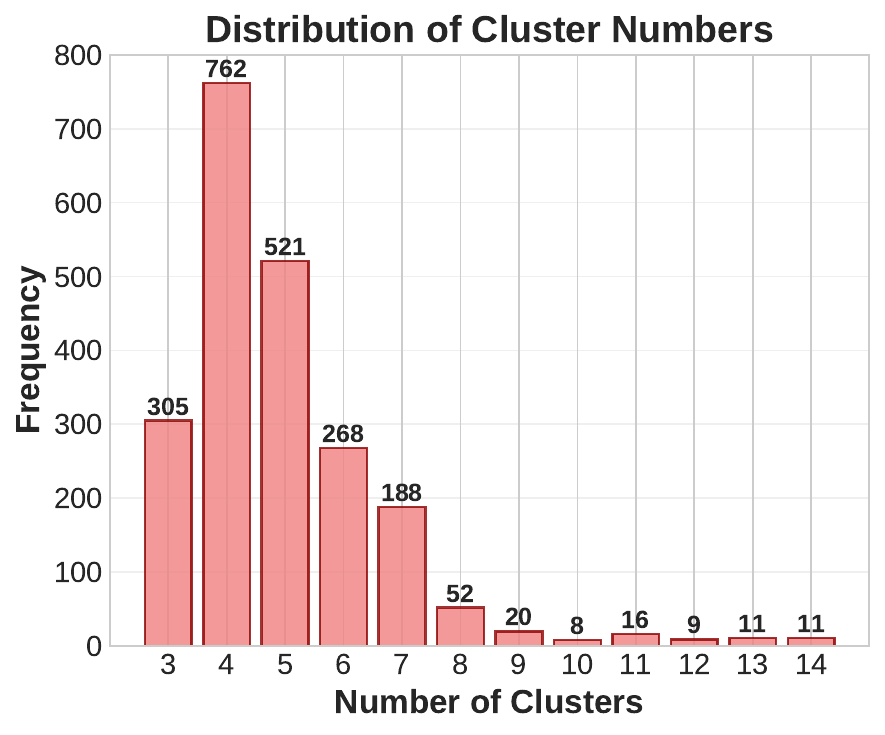}
    \caption{Cluster number distribution.}
    \label{fig:subfig-num}
\end{subfigure}
\hfill
\begin{subfigure}[t]{0.32\linewidth}
    \centering
    \includegraphics[width=\linewidth]{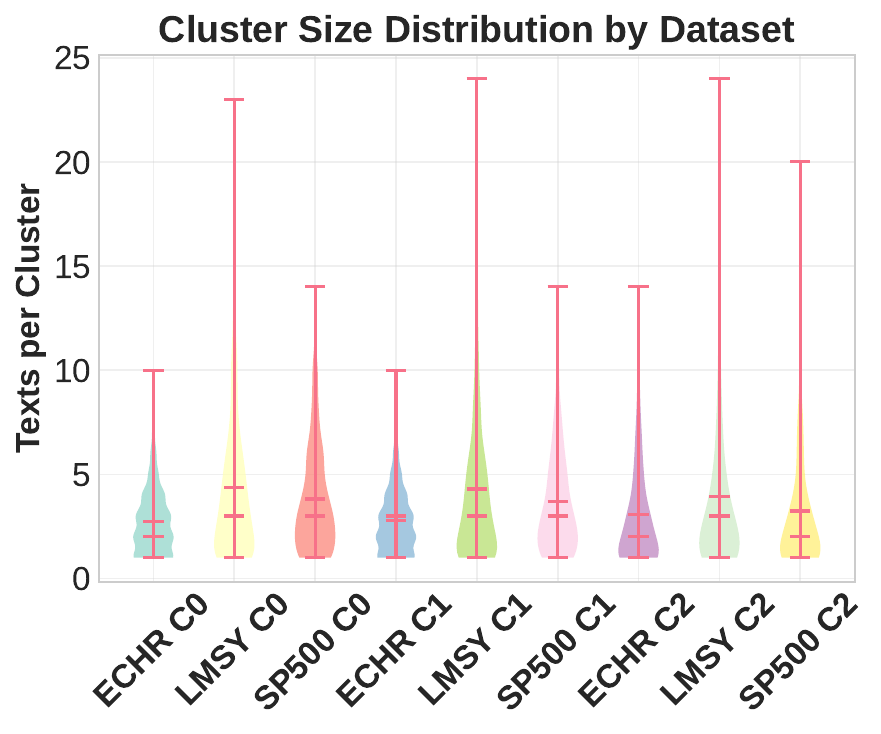}
    \caption{Cluster size by dataset.}
    \label{fig:subfig-size}
\end{subfigure}

\caption{Overview of the benchmark evaluation splits: (a) dataset statistics by source and split, where text and input lengths are measured in tokens, (b) distribution of the number of clusters per data example, and (c) distribution of cluster size, indicating the number of text instances per cluster for each dataset.}
%(a) dataset statistics by source and split, where text and input lengths are measured in tokens, (b) distribution of the number of clusters per data example, and (c) distribution of cluster size, indicating the number of text instances per cluster for each dataset.}

\vspace{-5pt}
\label{fig: benchmark-overview}

\end{figure*}

\subsection{Datasets}
\label{subsec: datasets}

We construct our benchmark using three complementary datasets that span conversational, legal, and financial domains, enabling evaluation of instruction-following clustering under diverse linguistic structures and reasoning demands.

\noindent\textbf{LMSYS-Chat.} Prior intent-discovery benchmarks such as Bank77 and CLINC~\citep{Casanueva2020bank77intent,larson-etal-2019-CLINICintent,fitzgerald2022massiveintent}  focus on short, single-turn utterances with narrowly defined intents, limiting their ability to capture the contextual and complex nature of real-world dialogue. We therefore use LMSYS-Chat, a large-scale corpus of real human--LLM conversations, to evaluate instruction-following clustering under realistic conversational settings.

\noindent\textbf{ECHR and S\&P 500.}
We include long-form documents from legal and financial domains. These texts are structurally complex and require substantive domain-specific reasoning, posing clustering challenges that differ fundamentally from dialogue. We aim to test whether models can group documents according to diverse abstract criteria beyond surface semantics. Detailed data preprocessing procedures, filtering criteria, and dataset statistics are provided in Appendix~\ref{appendix: data preprocess}.

\subsection{Benchmark Details}
\label{subsec: c0c1c2 construction}

Building on the synthesis pipeline in Section~\ref{subsec: data syn multi-agent}, we curate a benchmark that comprises 28 tasks, partitioned into 17 held-in tasks (available during training) and 11 held-out tasks (reserved exclusively for evaluation). Each task consists of one clustering instruction and a set of clusters, each populated with domain-specific texts. For all held-in tasks, we randomly split the texts within every cluster into two pools: 60\% of the texts are used to construct the training pool, while the remaining 40\% are reserved for in-domain evaluation.

% For every held-in task, we split the texts within each cluster 60/40: the 60\% portion forms the training pool, and the remaining 40\% is reserved for the C1 evaluation pool.
\noindent\textbf{Training set.} Training examples are generated by (i) randomly sampling a subset of clusters from a task and (ii) sampling multiple texts from the training pool of each selected cluster. 

\noindent\textbf{Evaluation splits.} We construct three evaluation splits that probe distinct generalization axes:
% \noindent\textbf{\dotsc C0} (\textcolor{ngreen}{seen tasks and seen texts}, \textcolor{nred}{unseen cluster combinations}): created by re-sampling new cluster–text combinations from the 60\% training pool that are not used to form training instances. 

% %Thus, both the instructions and the underlying texts were available during training, but the specific combinations are new.
% \noindent\textbf{C1} (\textcolor{ngreen}{seen tasks}, \textcolor{nred}{unseen texts and unseen cluster combinations}): constructed from the 40\% per-cluster reserved C1 evaluation pool. The clustering instructions are seen during training, but none of the texts in C1 appear in the training.

% \noindent\textbf{C2} (\textcolor{nred}{unseen tasks, unseen texts, and unseen cluster combinations}): drawn exclusively from the 11 held-out tasks whose clustering instructions and texts never appear during training.

\begin{itemize}[leftmargin=*, itemsep=-2pt, topsep=1pt,]
\item \textbf{C0} (\textcolor{ngreen}{seen tasks and seen texts}, \textcolor{nred}{unseen cluster combinations}): created by re-sampling a new subset of clusters and texts from the 60\% training pool that are not used to form training instances. 

%Thus, both the instructions and the underlying texts were available during training, but the specific combinations are new.
\item \textbf{C1} (\textcolor{ngreen}{seen tasks}, \textcolor{nred}{unseen texts and unseen cluster combinations}): constructed from the 40\% per-cluster reserved evaluation pool. The clustering instructions are seen during training, but none of the texts in the evaluation pool appear in the training.
\item \textbf{C2} (\textcolor{nred}{unseen tasks, unseen texts, and unseen cluster combinations}): drawn exclusively from the 11 held-out tasks whose clustering instructions, clusters, and texts never appear during training.
\end{itemize}

This protocol isolates different generalization axes: C0 tests recombination robustness with familiar instructions and texts, C1 tests transfer to new texts under familiar instructions, and C2 tests transfer to entirely new instructions and texts. The statistics of our benchmark are shown in Figure~\ref{fig: benchmark-overview}. Details of clustering instructions are provided in Appendix~\ref{appen: benchmark details}.

\section{Cluster-R1}
\label{sec: methods}

\subsection{Reasoning Distillation for Instruction-Following Clustering}
\label{sec:distill}

Instruction-tuned LLMs can be prompted for clustering but often fail to reliably decompose complex instructions~\citep{chen2025rmr1}. To address this, we first conduct reasoning distillation, which distills step-by-step clustering rationales from an oracle model to the student model. We first sample $M$ examples from the original corpus $\mathcal{D}$, yielding $\mathcal{D}_{\text{sub}} = \{(x^{(i)}, C^{(i)})\}_{i=1}^M$, where $x^{(i)}$ is an instruction-annotated input and $C^{(i)} = \{C^{(i)}_1,\dots,C^{(i)}_{k^{(i)}}\}$ is the gold clustering. An oracle model is queried with $(x^{(i)}, C^{(i)})$ to produce a reasoning trace $r^{(i)}$ that (1) infers cluster count $k^{(i)}$, (2) explains grouping logic, and (3) provides coherent cluster labels. This yields a distillation dataset:
\begin{equation}
    \mathcal{D}_{\text{distill}} = \{(x^{(i)}, y^{(i)}_{\text{trace}})\}_{i=1}^M,
\end{equation}
used for supervised fine-tuning. The training objective is the standard autoregressive loss:
\begin{equation}
\mathcal{L}_{\text{distill}}(\theta) = - \!\!\sum_{(x,y)\in\mathcal{D}_{\text{distill}}}\sum_{t=1}^{|y|}\! \log p_\theta(y_t \mid x, y_{<t}),
\end{equation}
encouraging structured reasoning aligned with clustering instructions. Further details on reasoning trace generation and refinement are provided in Appendix~\ref{appen:reasoning_chain_generation_detail}.

\subsection{Reinforcement Learning with Multiplicative Hybrid Rewards}
\label{sec:rl}

To further align model generation with the requirements of instruction-following clustering, we apply reinforcement learning to optimize clustering accuracy while preventing degenerate or structurally invalid outputs. We decompose the overall reward into a format-level constraint and a clustering-level objective:

% We refine the policy $\pi_\theta$ with GRPO, combining structural and clustering feedback via multiplicative hybrid rewards. For candidate outputs $\{y_j\}_{j=1}^G$, the total reward is
\begin{equation}
\mathcal{R}(y) = \mathcal{R}_{\text{format}}(y) + \mathcal{R}_{\text{clust}}(y).
\end{equation}

\paragraph{Format reward.} The format reward enforces strict syntactic and structural correctness of the generated outputs to avoid missing items, duplicate assignments, or malformed cluster declarations. Degenerated outputs are unusable regardless of semantic quality. We therefore define:
\begin{equation}
\mathcal{R}_{\text{format}}(y) = 
\begin{cases} 
1 & \text{format constraints satisfied}, \\[-2pt]
-1 & \text{otherwise}.
\end{cases}
\end{equation}
Further details of format constraints are provided in Appendix~\ref{appen:format_constraints}.

\paragraph{Clustering reward.}
Beyond structural validity, effective clustering requires both accurate inference of the number of clusters and high-quality cluster assignment. Accordingly, we design the clustering reward as a composition of \emph{count reward} and \emph{quality reward}:
\begin{equation}
\mathcal{R}_{\text{clust}}(y) = \big(\mathcal{R}_{\text{count}}(y) + \epsilon\big)\big(\mathcal{R}_{\text{qual}}(y) + \epsilon\big) - \epsilon,
\end{equation}
where $\epsilon = 0.1$ is a small constant introduced for numerical stability. The count reward $\mathcal{R}_{\text{count}}$ supervises the model’s ability to infer the number of clusters from the instruction and corpus. We define:
\begin{equation}
\mathcal{R}_{\text{count}}(y) = \gamma^{|\hat{K} - K|},
\end{equation}
where $\hat{K}$ and $K$ denote the predicted and ground-truth cluster counts, respectively, and $\gamma \in (0,1)$ is a decay factor (default $\gamma = 0.7$) that penalizes larger deviations more severely. The quality reward $\mathcal{R}_{\text{qual}}$ measures the semantic coherence of the predicted clustering using V-measure, a standard clustering metric that balances homogeneity and completeness (see Appendix~\ref{appen: vmeasure}):
\begin{equation}
\mathcal{R}_{\text{qual}}(y) = V(\hat{\mathcal{C}}, \mathcal{C}^*).
\end{equation}

% \paragraph{Clustering reward.} Beyond structural validity, effective clustering requires both accurate inference of the number of clusters and high-quality partition alignment with the latent ground truth. Accordingly, we design the clustering reward as a multiplicative composition of two complementary signals: a \emph{count reward} and a \emph{quality reward}:
% \begin{equation}
% \mathcal{R}_{\text{clust}}(y) = \big(\mathcal{R}_{\text{count}}(y)+\epsilon\big)\big(\mathcal{R}_{\text{qual}}(y)+\epsilon\big) - \epsilon,
% \end{equation}
% where $\epsilon = 0.1$ is a constant added for stability. 
% The count reward \mathcal{R}_{\text{clust}} explicitly supervises the model’s ability to infer the correct number of clusters from the instruction and corpus. Unlike traditional clustering pipelines where the cluster count is externally specified, our generative formulation requires the model to reason about granularity.
% :
% \begin{equation}
% \mathcal{R}_{\text{count}}(y) = \gamma^{|\hat{K} - K|},
% \end{equation}
% where $\hat{K}$ and $K$ denote the predicted and ground-truth cluster counts, respectively, and $\gamma \in (0,1)$ is a tunable decay parameter (default $\gamma = 0.7$). 

% The quality reward \mathcal{R}_{\text{qual} measures the semantic coherence of the predicted clusters using V-measure (details is in Appendix~\ref{appen: vmeasure}), which is the common pratice for evaluating text clustering thatbalances homogeneity and completeness,:
% \begin{equation}
% \mathcal{R}_{\text{qual}}(y) = V(\hat{\mathcal{C}}, \mathcal{C}^*).
% \end{equation}

\begin{table*}[ht]
\centering
\resizebox{2.0\columnwidth}{!}{%
\begin{tabular}{l*{13}{r}}
\toprule
\multirow{2}{*}{\textbf{Model}} 
& \multicolumn{3}{c}{\textbf{LMSY}} 
& \multicolumn{3}{c}{\textbf{ECHR}} 
& \multicolumn{3}{c}{\textbf{SP500}} 
& \multicolumn{3}{c}{\textbf{AVG}} 
& \textbf{OVERALL} \\
\cmidrule(lr){2-4}\cmidrule(lr){5-7}\cmidrule(lr){8-10}\cmidrule(lr){11-13}\cmidrule(lr){14-14}
& C0 & C1 & C2
& C0 & C1 & C2
& C0 & C1 & C2
& C0 & C1 & C2
& V \\
\midrule
\multicolumn{14}{l}{\textbf{\textit{Reasoning Models}}} \\
\midrule
\rowcolor{maroon!10} o3 & 
63.60 & 63.42 & \textbf{64.17} 
& \textbf{85.75} & \textbf{85.79} & 47.04 
& 57.01 & 59.04 & 66.51 
& 68.79 & 69.42 & 59.24 
& 65.08 \\
\rowcolor{maroon!10} Gemini 2.5 Pro & 
55.98 & 56.06 & 57.95 
& 85.05 & 84.14 & 64.29
& 57.60 & 63.43 & \textbf{69.34} 
& 66.21 & 67.88 & 60.92
& 61.82 \\
GPT-oss-120B & 
48.74 & 49.30 & 52.77 
& 70.54 & 69.18 & 52.58 
& 44.76 & 48.15 & 51.84 
& 54.68 & 55.54 & 52.40 
& 52.31 \\
DeepSeek-R1 & 
42.96 & 43.64 & 45.80 
& 53.78 & 54.55 & 34.06 
& 41.57 & 40.47 & 39.69 
& 46.10 & 46.22 & 39.85 
& 44.06 \\
Distill-Llama-70B & 
37.80 & 39.04 & 47.76
& 63.45 & 66.40 & 57.79
& 39.96 & 43.78 & 42.13
& 47.07 & 49.74 & 49.23 
& 45.12 \\
Distill-Qwen-32B & 
26.50 & 27.89 & 39.45
& 54.43 & 53.15 & 44.04
& 31.82 & 33.91 & 38.74
& 37.58 & 38.32 & 40.74 
& 34.96 \\
QwQ-32B & 
47.51 & 48.84 & 53.74
& 75.98 & 72.63 & 56.68
& 51.95 & 55.30 & 63.23
& 58.48 & 58.92 & 57.88 
& 54.78 \\
\midrule
\multicolumn{14}{l}{\textbf{\textit{General Models}}} \\
\midrule
\rowcolor{maroon!10} GPT-4o & 
26.95 & 30.15 & 33.38 
& 58.86 & 56.15 & 52.86 
& 40.53 & 40.26 & 31.15 
& 42.11 & 42.19 & 39.13 
& 41.26 \\
\rowcolor{maroon!10} GPT-4.1 & 
37.27 & 39.99 & 43.42 
& 64.93 & 59.87 & 40.05 
& 42.45 & 44.64 & 41.27 
& 48.22 & 48.17 & 41.25 
& 43.51 \\

Llama-3.1-70B-Instruct & 
24.57 & 27.50 & 28.97
& 42.69 & 44.58 & 35.77
& 36.21 & 37.66 & 32.28
& 34.49 & 36.58 & 32.34 
& 31.55 \\
Qwen2.5-72B-Instruct & 
18.85 & 20.08 & 27.78 
& 45.57 & 44.03 & 51.77 
& 35.34 & 36.17 & 33.28 
& 33.25 & 33.43 & 37.61 
& 29.06 \\
\midrule
\multicolumn{14}{l}{\textbf{\textit{Our Models}}} \\
\midrule
C1-Qwen-7B & 
65.63 & 65.04 & 58.51
& 81.81 & 82.76 & 52.66
& \textbf{69.35} & \textbf{68.08} & 60.28
& 72.26 & 71.96 & 57.15 
& 66.54 \\
C1-Qwen-14B & 
\textbf{66.90} & \textbf{66.47} & 62.65
& 84.05 & 84.80 & \textbf{66.39}
& 68.33 & 67.96 & 59.44
& \textbf{73.09} & \textbf{73.08} & \textbf{62.83} 
& \textbf{68.42} \\
\bottomrule
\end{tabular}%
}
\caption{V-measure (\%) score across three datasets under different evaluation splits (C0, C1, C2; see Section~\ref{subsec: c0c1c2 construction} for details). The final column presents the overall V-measure computed over all individual samples. Bold indicates the best result, and proprietary models are highlighted. Reasoning models outperform general models across all settings. Our models achieve the strongest overall performance, indicating that explicit reasoning supervision is crucial for faithful instruction-following clustering.}
\vspace{-10pt}
\label{tab:main-results-lrm-llm}
\end{table*}

\paragraph{Optimization.}
The GRPO optimization objective is defined as:
\begin{equation} \max_\theta \; \mathbb{E}_{y\sim \pi_\theta}[ \mathcal{R}(y)] - \beta \,\mathbb{D}_{\text{KL}}(\pi_\theta \,\|\, \pi_{\text{ref}}), \end{equation}
where $\pi_{\mathrm{ref}}$ denotes the distilled SFT checkpoint. To reduce variance during policy updates, group-relative advantages are computed for each candidate output:
\begin{equation}
A_j = \mathcal{R}(y_j) - \frac{1}{G}\sum_{k=1}^{G}\mathcal{R}(y_k),
\end{equation}
where $G$ is the number of samples in the group.

\section{Experimental Settings}

\paragraph{Baselines}
We compare our models with both LRMs and general LLMs, as well as embedding-based clustering approaches. For LRMs and LLMs, we consider both proprietary models (e.g., GPT-4o, Gemini-2.5-Pro) and state-of-the-art open-source models (e.g., DeepSeek-R1, Qwen2.5-72B-Instruct). For embedding-based methods, we evaluate general-purpose embeddings (e.g., OpenAI-Embedding\footnote{The latest OpenAI embedding model as of September 12, 2025: \href{https://platform.openai.com/docs/guides/embeddings}{\texttt{text-embedding-3-large}}}, all-MiniLM-L6-v2~\citep{wang2020minilm}) and instruction-tuned embedding models (e.g., InBedder~\citep{answer-need}, Instructor~\cite{instructor}), each combined with classical clustering algorithms such as K-means and GMMs. The implementation details for all the baselines are provided in Appendix~\ref{appen: embedding_methods}.

\paragraph{Implementation Details} 
We adopt the Qwen-2.5-Instruct series models~\citep{qwen25_techreport_2024} as the backbone due to their strong reasoning capabilities acquired during pre-training.
Comprehensive configuration and training datasets is provided in Appendix~\ref{appendix:implementation_details}.
For evaluation, we employ the V-measure metric, following established practice in clustering assessments~\citep{muennighoff2022mteb, vmeasure}. Formal description of V-measure is shown in Appendix~\ref{appen: vmeasure}.

% To operationalize this, we leverage an instruction-following embedding function \(\mathbf{Emb}(\cdot \mid I): \mathcal{X} \to \mathbb{R}^d\), which maps each input text \(\mathbf{x}_i\) to a \(d\)-dimensional embedding \(\mathbf{z}_i = \mathbf{Emb}(\mathbf{x}_i \mid I)\), conditioned on the instruction \(I\). The similarity between embeddings is typically measured using cosine similarity:
% \[
% \textrm{Sim}(\mathbf{z}_i, \mathbf{z}_j) = \frac{\mathbf{z}_i^\top \mathbf{z}_j}{\|\mathbf{z}_i\| \|\mathbf{z}_j\|}.
% \]
% A standard clustering algorithm, such as \(k\)-means, is then applied to the set of conditional embeddings \(\{ \mathbf{z}_i \}_{i=1}^n\) to obtain \(\mathcal{C}\):
% \[
% \mathcal{C} = \arg\min_{\mathcal{C}} \sum_{m=1}^k \sum_{\mathbf{z}_i \in C_m} \| \mathbf{z}_i - \boldsymbol{\mu}_m \|^2,
% \]
% where \(\boldsymbol{\mu}_m\) is the centroid of cluster \(C_m\).

% This formulation allows the clustering to adapt dynamically to different instructions without retraining the clustering algorithm itself, relying instead on the instruction-awareness of the embedding model. Evaluation of such clusterings can be performed using metrics like the V-measure or normalized mutual information, conditioned on ground-truth labels aligned with the instruction's intent.

\section{Main Results}
\label{sec: exeriments}
% This section evaluates StructRAG by conducting experiments on knowledge-intensive tasks.

\begin{table*}[h]
\centering
\resizebox{2.0\columnwidth}{!}{%
\begin{tabular}{l*{13}{r}}
\toprule
\multirow{2}{*}{\textbf{Model}} 
& \multicolumn{3}{c}{\textbf{LMSY}} 
& \multicolumn{3}{c}{\textbf{ECHR}} 
& \multicolumn{3}{c}{\textbf{SP500}} 
& \multicolumn{3}{c}{\textbf{AVG}} 
& \textbf{OVERALL} \\
\cmidrule(lr){2-4}\cmidrule(lr){5-7}\cmidrule(lr){8-10}\cmidrule(lr){11-13}\cmidrule(lr){14-14}
& C0 & C1 & C2
& C0 & C1 & C2
& C0 & C1 & C2
& C0 & C1 & C2
& V \\
\midrule
\multicolumn{14}{l}{\textbf{\textit{K-means}}} \\
\midrule
\rowcolor{maroon!10} Qwen3-Embedding & 55.88 & 56.28 & 53.26 & 74.58 & 71.87 & 59.45 & 53.41 & 54.67 & 60.81 & 61.29 & 60.94 & 57.84 & 58.31 \\
\rowcolor{maroon!10} InBedder & 52.19 & 52.05 & 45.95 & 67.46 & 66.08 & 48.13 & 53.06 & 53.97 & 49.72 & 57.57 & 57.37 & 47.93 & 53.28 \\
\rowcolor{maroon!10} Instructor & 47.20 & 47.04 & 44.04 & 58.12 & 57.42 & 47.30 & 48.74 & 49.12 & 47.28 & 51.35 & 51.19 & 46.21 & 48.55 \\
OpenAI-Embedding & 50.63 & 51.28 & 49.11 & 56.58 & 55.99 & 45.75 & 47.74 & 48.82 & 48.47 & 51.65 & 52.03 & 47.78 & 50.66 \\
all-MiniLM-L6-v2 & 44.39 & 45.07 & 43.27 & 58.00 & 57.58 & 49.07 & 48.44 & 48.78 & 48.75 & 50.27 & 50.48 & 47.03 & 47.51 \\

\midrule
\multicolumn{14}{l}{\textbf{\textit{Gaussian Mixture}}} \\
\midrule
\rowcolor{maroon!10} Qwen3-Embedding & 55.64 & 56.13 & 52.23 & 72.22 & 71.64 & 59.23 & 54.12 & 54.53 & 59.77 & 60.66 & 60.77 & 57.08 & 57.88 \\
\rowcolor{maroon!10} InBedder & 49.02 & 49.40 & 42.90 & 63.95 & 64.49 & 46.53 & 48.82 & 49.85 & 46.36 & 53.93 & 54.58 & 45.26 & 50.26 \\
\rowcolor{maroon!10} Instructor & 42.89 & 43.39 & 41.52 & 55.75 & 55.16 & 44.51 & 44.55 & 46.46 & 44.80 & 47.73 & 48.34 & 43.61 & 45.22 \\
OpenAI-Embedding & 46.11 & 47.68 & 44.98 & 55.84 & 55.46 & 44.12 & 44.83 & 45.94 & 46.84 & 48.93 & 49.69 & 45.32 & 47.54 \\
all-MiniLM-L6-v2 & 40.80 & 41.48 & 40.09 & 54.01 & 54.05 & 44.57 & 43.67 & 44.76 & 47.51 & 46.16 & 46.76 & 44.06 & 43.91 \\
\midrule
\multicolumn{14}{l}{\textbf{\textit{Our Models}}} \\
\midrule
C1-Qwen-7B & 
65.63 & 65.04 & 58.51
& 81.81 & 82.76 & 52.66
& \textbf{69.35} & \textbf{68.08} & \textbf{60.28}
& 72.26 & 71.96 & 57.15 
& 66.54 \\
C1-Qwen-14B & 
\textbf{66.90} & \textbf{66.47} & \textbf{62.65}
& \textbf{84.05} & \textbf{84.80} & \textbf{66.39}
& 68.33 & 67.96 & 59.44
& \textbf{73.09} & \textbf{73.08} & \textbf{62.83} 
& \textbf{68.42} \\
\bottomrule
\end{tabular}%
}
\caption{Comparison of embedding-based clustering against our models. Results of instruction-following embedders (Qwen3-Embedding, InBedder, and Instructor) are highlighted. Bold indicates the
best result. While instruction-tuned embedders outperform generic embedders, they remain consistently inferior to reasoning-based clustering.\looseness-1}
\label{tab:embedding-vs-ours}
\end{table*}

% \subsection{Main Results}
\label{subsec: main_exeriments_results}

\noindent\textbf{C1-Qwen achieves the best overall performance.}
Table~\ref{tab:main-results-lrm-llm} and Table~\ref{tab:embedding-vs-ours} show that our models consistently outperform all baselines across settings, with the 14B variant achieving the highest average score of 68.42\%. These results demonstrate the effectiveness of our training framework for reasoning-intensive, instruction-following clustering. Case studies are provided in Appendix~\ref{appen: analyze_main_results}.

\noindent\textbf{Reasoning models substantially outperform general-purpose LLMs.}
As shown in Table~\ref{tab:main-results-lrm-llm}, general-purpose models such as GPT-4o and GPT-4.1 achieve only 41--44\% on average, whereas reasoning-oriented models reach 55--65\%. This gap primarily arises because general LLMs frequently violate clustering constraints, producing duplicate items or structurally invalid outputs. Nevertheless, high format accuracy alone does not guarantee superior clustering performance. Models such as Llama-3.1-70B-Instruct exhibit high format accuracy but still attain low V-measure scores. Detailed format error analysis is presented in Figure~\ref{fig: grid}.

\noindent\textbf{Instruction-following embedders improve over generic embeddings but lag behind reasoning models.}
Table~\ref{tab:embedding-vs-ours} compares our models with embedding-based approaches. Instruction-following embedders (e.g., InBedder and Instructor) outperform generic embedding models because they concatenate language instructions with text data during training, yet they still fall short of reasoning models. The strongest embedding-based baseline (Qwen3-Embedding + K-means) achieves 58.3\% on average, nearly 10 points below our 14B model, highlighting the importance of explicit reasoning for instruction-driven clustering.

\begin{table}[h]
    \begin{center}
        \resizebox{1.0\columnwidth}{!}{
        \begin{tabular}{@{}lcccc@{}}
            \toprule
            \textbf{Method} 
            & \textbf{C0} 
            & \textbf{C1} 
            & \textbf{C2} 
            & \textbf{Ave.} \\
            \midrule
            SFT + Final Answer 
            & 65.51 & 52.70 & 8.20 & 42.14 \\
            SFT + Distilled + Final Answer 
            & 64.87 & 54.76 & 11.98 & 43.87 \\
            SFT + Distilled + RL & 72.26 & 71.96 & 57.15 & 67.12 \\
            \bottomrule
        \end{tabular}
            }
    \end{center}
    \caption{Comparison of reasoning training versus SFT directly on final answer. Reasoning supervision is essential for generalization: models trained without reasoning fail on C2, while the full pipeline remains robust.     
    %Models trained directly on final answers fail to generalize, particularly on unseen tasks (C2). Incorporating reasoning distillation followed by RL yields consistent gains, confirming that reasoning supervision are essential for robust clustering.
    % Scores for C0, C1, and C2 are averages over all corresponding samples across three datasets. The Ave. score reports the overall mean computed across all individual samples.
    }
    \vspace{-3mm}
    \label{tab: comparison_with_sft}
\end{table}

% \begin{figure*}[h]
%     \centering
%     % Table in the first minipage
%     \begin{minipage}[t]{0.6\linewidth}
%         \vspace*{5pt} % Anchor for top alignment
%         \centering
%         \scalebox{0.7}{
%         \begin{tabular}{@{}lcccc@{}}
%             \toprule
%             \textbf{Method} 
%             & \textbf{C0} 
%             & \textbf{C1} 
%             & \textbf{C2} 
%             & \textbf{Ave.} \\
%             \midrule
%             SFT + Final Answer 
%             & 65.51 & 52.70 & 8.20 & 42.14 \\
%             SFT + Distilled + Final Answer 
%             & 64.87 & 54.76 & 11.98 & 43.87 \\
%             SFT + Distilled + RL & 72.26 & 71.96 & 57.15 & 67.12 \\
%             \bottomrule
%         \end{tabular}
%         }
%         % Use 'captionof' for a table inside a figure environment
%         \captionof{table}{Comparison of reasoning-based training versus SFT directly on final answer.}
%         \label{tab: comparison_with_sft}
%     \end{minipage}% <--- The '%' is important to prevent unwanted space
%     \hfill
%     % Figure in the second minipage
%     \begin{minipage}[t]{0.38\linewidth}
%         \vspace*{-15pt} % Anchor for top alignment
%         \centering
%         \includegraphics[width=0.8\linewidth]{figures/Model Sizel Scaling.pdf}
%         \vspace{-12pt}
%         \caption{Model Size Scaling}
%         \label{fig:model_size_scaling}
%     \end{minipage}
% \end{figure*}

\section{Analysis}
This section presents empirical analyses to identify the key components required to train effective clustering agents. Our study covers impact of reasoning training, training strategies, and scaling effects.

\subsection{Effectiveness of Reasoning Training}
\label{subsec:reasoning_training}

% These results underscore two key insights. First, reasoning supervision improves not only in-domain 
% clustering quality but also generalization to novel instructions and domains. Second, reinforcement 
% learning further strengthens reasoning fidelity by aligning generated traces with task-specific 
% objectives. Taken together, the evidence demonstrates that reasoning is not merely an auxiliary 
% artifact but a crucial training signal that drives instruction-faithful clustering.

We analyze the effectiveness of reasoning-based training and demonstrate its superiority over answer-only approaches. Specifically, we compare:

\noindent\textbf{(1) SFT + Final Answer:} Fine-tunes the instruction model to predict the correct final answer directly, without intermediate reasoning chains.  

\noindent\textbf{(2) SFT + Distilled + RL:} Our proposed method, which first applies SFT on distilled reasoning data and then incorporates RL, as detailed in Section~\ref{sec: methods}.  

\noindent\textbf{(3) SFT + Distilled + Final Answer:} Starts from the same distilled checkpoint as our method but is fine-tuned exclusively on final answers, isolating the effect of RL in eliciting reasoning.

\noindent\textbf{Reasoning supervision (distillation + RL) substantially enhances clustering capability and generalizability}: Table~\ref{tab: comparison_with_sft} shows the combination of reasoning supervision and RL results in significant improvements. The baseline SFT + Final Answer achieves an average V-measure of 42.14 and 8.20 on the C2 split, indicating that the model tends to memorize surface patterns rather than develop a transferable reasoning capability. Adding reasoning distillation improves average performance modestly to 43.87, indicating that exposure to explicit reasoning traces helps internalize clustering logic but remains insufficient for robust generalization. In contrast, Distilled + RL pipeline boosts the average V-measure to 67.12. Overall, these results demonstrate that reasoning supervision (distillation + RL) enhances clustering competence for LRMs.

% These results highlight that reinforcement learning with hybrid rewards is essential for eliciting coherent reasoning behavior: it encourages adherence to format constraints, accurate cluster-count inference, and semantically faithful grouping. 

% \begin{table}[h]
%     \centering
%     \scalebox{0.85}{
%     \begin{tabular}{@{}lcccc@{}}
%         \toprule
%         \textbf{Method} 
%         & \textbf{C0 (V)} 
%         & \textbf{C1 (V)} 
%         & \textbf{C2 (V)} 
%         & \textbf{Average (V)} \\
%         \midrule
%         \multicolumn{5}{l}{\textbf{Train on 9k (Distillation) Data}} \\
%         \midrule
%         SFT + Final Answer 
%         &  &  &  &  \\
%         SFT + \textbf{Distilled} 
%         &  &  &  &  \\
%         \midrule
%         \multicolumn{5}{l}{\textbf{Train on Full Data}} \\
%         \midrule
%         SFT + Final Answer 
%         & 65.51 & 52.70 & 8.20 & 42.14 \\
%         SFT + Distilled + Final Answer 
%         & 64.87 & 54.76 & 11.98 & 43.87 \\
%         SFT + Distilled + RL (\ours) 
%         &  &  &  &  \\
%         \bottomrule
%     \end{tabular}
%     }
%     \caption{\textbf{Comparison of reasoning-based training versus SFT across evaluation clusters C0–C2.} Metric is V-measure (\%). The final column shows average performance. * indicates reasoning-based methods.}
%     \label{tab:comparison_with_sft_c0c1c2}
% \end{table}

\subsection{Training Recipes for Eliciting Effective Reasoning}
\label{subsec:training_recipes}

% To identify strategies for eliciting effective reasoning in instruction-following clustering, we investigate the key components required to train reasoning-enabled clustering models. 
This section investigate the key steps for eliciting effective reasoning in instruction-following clustering. Specifically, we compare:

\noindent\textbf{(1) RL:} Applies cold-start RL with rule-based rewards focused on cluster correctness and format compliance (see Section~\ref{sec: methods}). The model learns solely from reward feedback without any prior reasoning guidance.  

\noindent\textbf{(2) RL + Reasoning Hint:} Extends the baseline RL setup by modifying the prompt to include explicit reasoning hints. The prompt first instructs the model to infer latent clusters and then assign items accordingly, encouraging step-by-step reasoning rather than direct prediction. The reasoning hint prompt remain consistent with the distillation prompt (see the think process in Figure~\ref{fig: reason_chain_gen}).

\noindent\textbf{(3) SFT + RL + Reasoning Hint:} Builds on the previous setup with an additional distillation stage from stronger teacher models as a warm start before RL training. With RL alone, weaker models (especially at smaller scales) often fail to explore high-quality reasoning chains for clustering. Distilling strong reasoning traces on a small subset of data mitigates this limitation.

\begin{figure}[ht]
    \centering
    \includegraphics[width=1.0\linewidth]{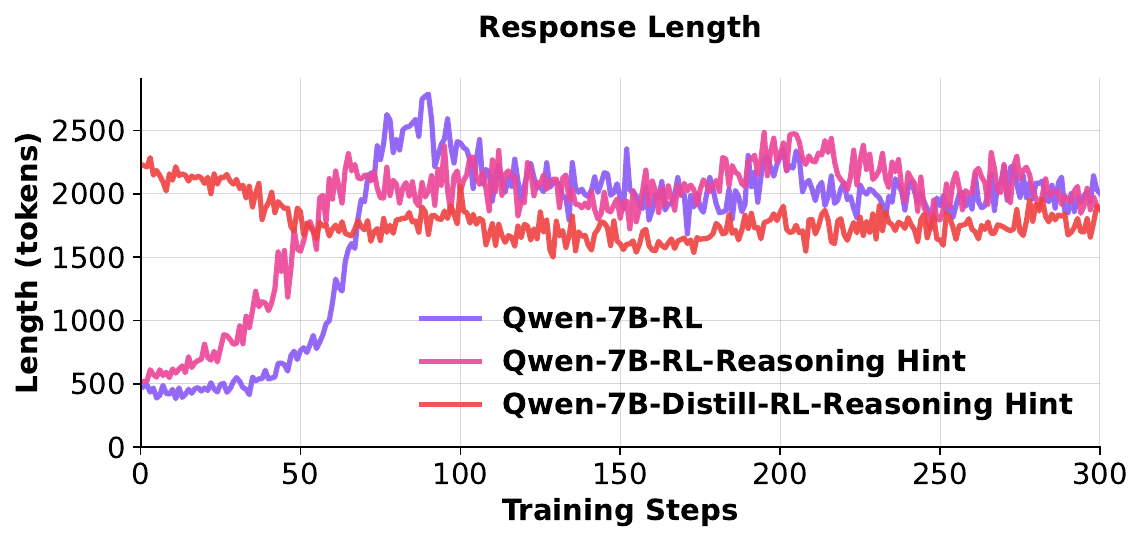}
    \vspace{-4pt}
    \includegraphics[width=1.0\linewidth]{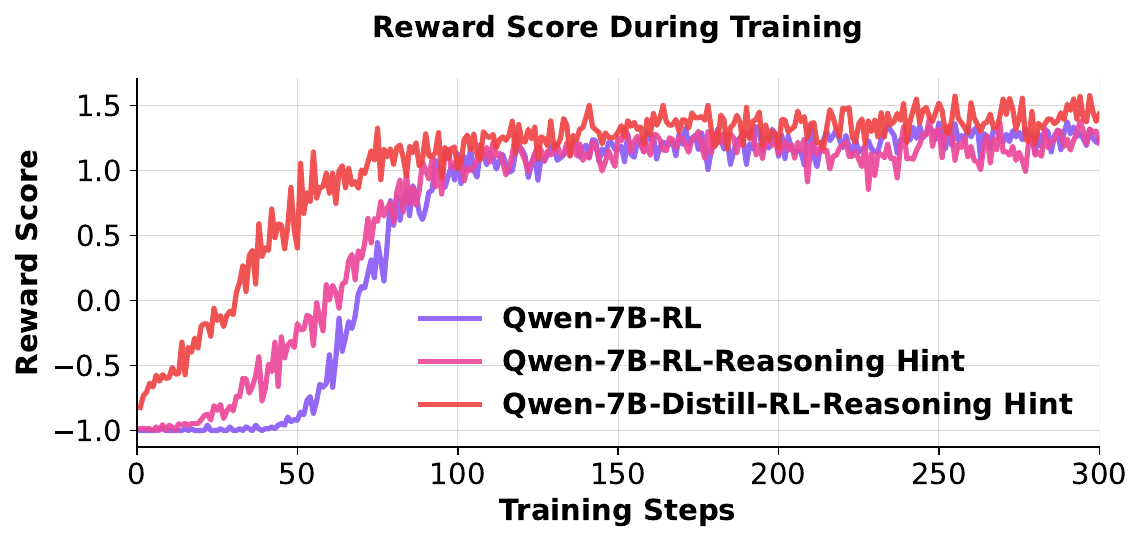}
    \caption{Training dynamics of different training recipes for the Qwen-7B model. Models initialized with distilled reasoning traces converge faster, exhibit more stable response lengths, and achieve higher rewards.}
    \label{fig:training_dynamics_7b}
    %\vspace{-3mm}
\end{figure}

\noindent\textbf{SFT + RL + Reasoning Hint achieves the best performance}: Table~\ref{tab:training_recipes_doubled} shows that adding reasoning hints to the prompt improves clustering performance, particularly for stronger base models. The 14B model exhibits a clear gain over the 7B counterpart, indicating that larger models are more capable of internalizing and leveraging reasoning cues during RL. This suggests that explicit prompting can partially elicit reasoning behavior, but its effectiveness depends on model capacity. The SFT + RL + Reasoning Hint variant achieves the best overall performance. SFT provides a structured prior that stabilizes RL training and promotes consistent reasoning patterns. As shown in Figure~\ref{fig:training_dynamics_7b}, this approach yields more stable response lengths, faster convergence, and consistently higher reward scores. In summary, reasoning hints can trigger emergent reasoning in sufficiently capable models, while combining SFT with RL offers a more reliable and scalable pathway to reasoning-driven clustering.

\begin{table}[t]
    \centering
    \scalebox{0.75}{
    \begin{tabular}{@{}lcccc@{}}
        \toprule
        \textbf{Method} & \textbf{C0} & \textbf{C1} & \textbf{C2} & \textbf{Ave.} \\
        \midrule
        \multicolumn{5}{l}{\textbf{7B Base Model}} \\
        \midrule
        + RL & 46.72 & 48.19 & 46.52 & 44.90 \\
        + RL + Reasoning Hint & 48.19 & 49.99 & 46.07 & 46.42 \\
        + Distill + RL + Reasoning Hint & 66.05 & 65.97 & 56.46 & 61.80 \\
        \midrule
        \multicolumn{5}{l}{\textbf{14B Base Model}} \\
        \midrule
        + RL & 56.39 & 53.57 & 53.71 & 52.76 \\
        + RL + Reasoning Hint & 66.05 & 66.40 & 60.87 & 62.41 \\
        + Distill + RL + Reasoning Hint & 68.84 & 69.21 & 59.71 & 64.43 \\
        \bottomrule
    \end{tabular}
    }
    \caption{Comparison of training recipes across 7B and 14B models. Prompt-level reasoning hints offer limited benefits, whereas combining reasoning distillation with RL yields the strongest performance.}
    % Reasoning hints alone provide limited improvements, especially for smaller models. In contrast, combining reasoning distillation with RL consistently yields the strongest performance, indicating that structured reasoning priors are crucial for stable and effective policy optimization.
    \vspace{-3mm}
    \label{tab:training_recipes_doubled}
\end{table}

\subsection{Effect of Scaling Model size }
%\label{subsec: scaling model sie}

\begin{figure}[h]
    \centering
    \includegraphics[scale=0.55]{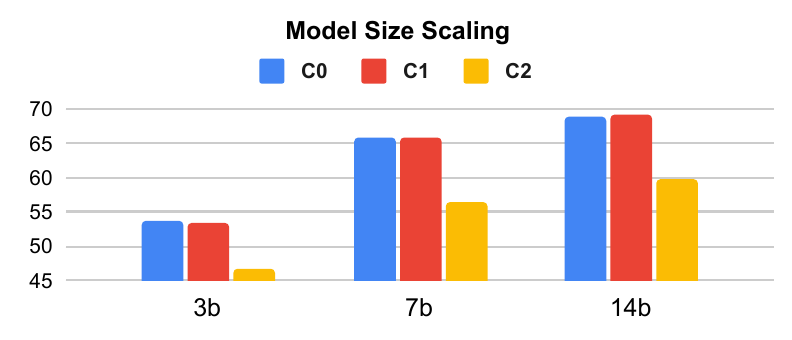}
    \caption{Effect of model size on clustering performance and generalization. The y-axis shows the average V-measure over all corresponding samples.}
    \vspace{-2mm}
    \label{fig:model_size_scaling}
\end{figure}

We investigate how instruction-following clustering performance scales with model capacity. Using the Qwen-2.5-Instruct model series, we compare models with 3B, 7B, and 14B parameters. All models are trained using our full two-stage pipeline (Section~\ref{sec: methods}). As illustrated in Figure~\ref{fig:model_size_scaling}, performance scales smoothly with model size across all evaluation conditions (C1 to C3). Larger models exhibit both higher average V-measures and smaller performance gaps between seen and unseen clustering tasks, indicating enhanced generalization and reasoning robustness.\looseness-1

% ACL requires a Limitations section
\vspace{1pt}
\section{Conclusion}

% This work presents a new paradigm for instruction-following clustering, reframing it as an end-to-end generative reasoning process. By leveraging LRMs trained through reasoning distillation and reinforcement learning, our approach unifies the interpretation of clustering instructions with the autonomous discovery of latent cluster structures and . Through extensive experiments on our \textsc{ReasonCluster}, a comprehensive benchmark spanning multiple domains and clustering intents, we demonstrate that explicit reasoning supervision leads to consistent gains in accuracy, interpretability, and generalization over both embedding-based and LRM baselines. By bridging explicit reasoning and representation learning, this framework paves the way toward more interpretable, autonomous, and instruction-aligned clustering systems.

This work introduces a new paradigm for instruction-following clustering by reframing it as an end-to-end generative reasoning task. By training LRMs with reasoning distillation and RL, our approach jointly interprets clustering instructions, infers the instruction-aligned cluster structure, and produces full cluster assignments within a single generative process, removing the need for external clustering heuristics. Extensive experiments on \textsc{ReasonCluster}, a novel and comprehensive benchmark spanning multiple domains and clustering intents, demonstrate that explicit reasoning supervision yields consistent improvements in clustering quality, interpretability, and generalization over both embedding-based methods and strong LRM baselines. By bridging explicit reasoning and representation learning, this framework paves the way toward more interpretable, autonomous, and instruction-aligned clustering systems.

% We envision future work extending this framework toward multi-turn reasoning, multi-modal clustering, and self-improving reasoning feedback loops, advancing autonomous models that can organize complex information in alignment with open-ended human instructions.

\clearpage
\vspace{-5pt}
\section{Limitations}
Despite the demonstrated effectiveness of our approach, two key limitations remain. The first lies in the context-length bottleneck of existing large reasoning models, which limits their ability to reason jointly over extensive corpora. This constraint hinders scalability to web-scale or streaming data. The second concerns format robustness: although reasoning distillation and reinforcement learning improve coherence, the model occasionally violates output constraints, producing duplicate or missing cluster assignments, particularly as input length increases. Such errors likely arise from attention degradation and the lost-in-the-middle effect inherent to long-context generation. To address these issues, future research will investigate memory-augmented or retrieval-enhanced clustering, as well as multi-agent or hierarchical clustering systems, enabling scalable, context-aware, and format-consistent instruction-following clustering.
\section*{Ethics Statement}
This work does not introduce new ethical risks beyond those commonly associated with large language models. All datasets used in this study are derived from publicly available or appropriately licensed sources and are processed solely for research purposes. We acknowledge the potential for downstream misuse of clustering systems and emphasize that our benchmark and models are intended for research evaluation rather than direct real-world deployment.
% ---------- References ----------
% Use ACL natbib style implicitly; DO NOT set \bibliographystyle here.
% Put your entries in, e.g., custom.bib (and/or anthology.bib).
\bibliography{custom} % or {anthology,custom}

% ---------- Appendix ----------
\clearpage
\appendix
% If you keep an appendix ToC, etoc/minitoc generally work, but many ACL tracks discourage extra ToC space.
% Comment these out if space is tight or if the CFP forbids them.
% \usepackage{etoc}
% \section*{Appendix Table of Contents}
% \startcontents[appendix]
% \printcontents[appendix]{}{0}{\large}

\section{Related Work}
\label{appen: related}

\paragraph{Traditional clustering.}
Classical clustering methods have long been foundational in natural language processing and data analysis.
Partition-based algorithms such as K-means \citep{macqueen1967some, lloyd1982least} and Gaussian mixture models \citep{dempster1977maximum} rely on distance or probabilistic assumptions to partition data into groups.
Hierarchical approaches \citep{johnson1967hierarchical, blashfield1978literature} construct tree-like taxonomies of nested relationships, while density-based methods such as DBSCAN \citep{ester1996density} identify clusters of arbitrary shapes without requiring the number of clusters in advance.
Spectral clustering \citep{ng2001spectral} leverages eigen-decomposition of similarity graphs to capture complex structures.
Additionally, topic modeling approaches—including probabilistic latent semantic analysis \citep{hofmann2001unsupervised} and latent Dirichlet allocation (LDA) \citep{blei2003latent}—have been widely used to uncover latent semantic topics.
In the neural era, deep clustering methods \citep{xie2016unsupervised, yang2016joint, huang2014deep, zhou2019end, zhang2021supporting} combine representation learning with clustering objectives, yielding substantial improvements over static pipelines.

\paragraph{Instruction-following embedders.}
Recent work highlights the limitations of general-purpose embeddings (e.g., Sentence-BERT \citep{reimers-gurevych-2019-sentencebert}, MiniLM \citep{wang2020minilm}, Qwen Embeddings \citep{qwen3-embedding}) when clustering tasks must adapt to user-specific objectives rather than generic semantic similarity.
To address this gap, instruction-following embedders condition embeddings on natural language prompts.
For example, Instructor \citep{instructor} and InBedder \citep{answer-need} finetune encoders to explicitly follow task instructions, improving controllability across diverse downstream applications.
While effective at aligning embeddings with user intent, these methods still treat clustering as a downstream algorithmic step, requiring separate unsupervised methods such as K-means or GMMs for grouping.
Consequently, they lack the ability to autonomously infer the number of clusters or reason about structural properties of the corpus \citep{zeng2025evaltree}.

\paragraph{LLM-involved clustering.}
Beyond embeddings, large language models (LLMs) and large reasoning models (LRMs) are increasingly used as active participants in clustering.
Instead of the ``embed-then-cluster'' paradigm, these models unify instruction following with reasoning over latent corpus structure.
\citet{zhang-etal-2023-clusterllm} proposed ClusterLLM, which leverages interactive triplet and pairwise feedback for refining clusters.
\citet{viswanathan2024large} showed that LLMs can augment document representations and generate pseudo-constraints for semi-supervised clustering.
Instruction-guided approaches also employ LLMs to clarify boundaries via natural language explanations \citep{wang2023goal, de2023idas}, or to iteratively refine edge cases for more coherent clusters \citep{feng2024llmedgerefine}.
More recently, reasoning-focused LRMs have been trained with hybrid supervised distillation and reinforcement learning to produce interpretable clustering rationales, faithful taxonomies, and stable cluster counts. These approaches represent a shift toward generation-based clustering that tightly integrates user instructions with autonomous structural discovery.

\section{Full Prompts}
\label{appendix: Prompt used for data synthesis}

The multi-agent data synthesis pipeline (Section~\ref{sec: bench construction}) employs structured prompting to ensure consistency and interpretability across stages. Specifically, Figure~\ref{fig:prompt_inst_taxonomy} shows the taxonomy generation expert prompt used to design reasoning-based clustering dimensions, Figure~\ref{fig:prompt_taxonomy_refine} illustrates the refinement prompt enforcing clarity and consistency, and Figure~\ref{fig:prompt_classification} presents the classification agent prompt for assigning categories under the refined taxonomy.

\begin{figure*}[t]
  \centering
  \begin{tcolorbox}[
      colback=gray!5!white,
      colframe=dartgreen,
      title=Cluster Generation Prompt,
      boxrule=0.3mm,
      arc=3mm,
      width=2.0\columnwidth
    ]
\small
You are an expert in taxonomy design.  
Given (i) a description of a dataset and (ii) example text entries from it, your task is to propose meaningful clustering dimensions that capture distinct reasoning-based perspectives.  

For each clustering dimension:  
1. Provide a concise description of what the dimension represents.  
2. List the possible cluster labels under this dimension.  
3. Justify why each cluster is distinct, highlighting reasoning factors such as cause, intent, context, or outcome.  

(i) \{Dataset description\}  
(ii) \{Datapoint examples\}  

Your output must follow this structured table format:  

\begin{center}
\begin{tabularx}{\columnwidth}{|c|l|X|}

id & name       & description \\

1  & Category A & what it is + why it's distinct \\
2  & Category B & what it is + why it's distinct \\
3  & Category C & what it is + why it's distinct \\

\end{tabularx}
\end{center}

  \end{tcolorbox}
  \caption{The system prompt of cluster generation agent.}
  \label{fig:prompt_inst_taxonomy}
\end{figure*}

\begin{figure*}[t]
  \centering
  \begin{tcolorbox}[
      colback=gray!5!white,
      colframe=dartgreen,
      title=Taxonomy Refinement Expert Prompt,
      boxrule=0.3mm,
      arc=3mm,
      width=2.0\columnwidth
    ]
\small
You are a taxonomy refinement expert. Your goal is to revise the given taxonomy so that it is clear, consistent, and practically usable. Follow the requirements below:

\textbf{Requirements}
\begin{itemize}
  \item \textbf{Category name}: $\leq$ \{name\_length\} words; concise, specific, and informative.
  \item \textbf{Description}: $\leq$ \{desc\_length\} words; clearly explain what distinguishes this category.
  \item \textbf{Mutual exclusivity}: Categories must not overlap or contradict each other.
  \item \textbf{Collective exhaustiveness}: Categories together should cover all possible intents in the given context.
  \item \textbf{Granularity}: All categories must be defined at the same level of specificity.
  \item \textbf{No vague labels}: Avoid terms like ``Other,'' ``General,'' or ``Miscellaneous.''
\end{itemize}

\textbf{Tasks}
\begin{enumerate}
  \item Review the existing taxonomy and suggest improvements (e.g., renaming, merging, splitting, or adding new categories).
  \item Ensure each category has a clear reasoning justification and aligns with the data context.
  \item If categories are missing, add enough to make the taxonomy collectively exhaustive.
\end{enumerate}

\textbf{Clustering Instruction}  
Based on the refined taxonomy, check the instruction and provide a \emph{neutral clustering instruction} that accurately reflects the categorization principle without revealing category labels. Do not include or hint at the actual category names in the instruction.

\vspace{1ex}

  \end{tcolorbox}
  \caption{The system prompt for taxonomy refinement and clustering instruction design.}
  \label{fig:prompt_taxonomy_refine}
\end{figure*}

\begin{figure*}[ht]
  \centering
  \begin{tcolorbox}[
      colback=gray!5!white,
      colframe=dartgreen,
      title=Expert Classification Agent Prompt,
      boxrule=0.3mm,
      arc=3mm,
      width=2.0\columnwidth
    ]
\small
You are an expert classification agent. Your task is to analyze the given data item and assign it to the most appropriate category from the provided taxonomy.  

\textbf{Taxonomy}  
\{taxonomy\}  

\textbf{Data to Classify}  
\{data\_item\}  

\textbf{Classification Instructions}
\begin{enumerate}
  \item Carefully read the data item.
  \item Compare it against the taxonomy categories and their descriptions.
  \item Identify the \textbf{single best-fitting category} that captures the primary characteristic of the item.
  \item If the item clearly does not fit any category, or if it is ambiguous, respond with \texttt{no\_match}.
\end{enumerate}

\textbf{Key Guidelines}
\begin{itemize}
  \item Focus only on the essential content of the data item.
  \item Use the category descriptions (especially ``distinct because'') to guide differentiation.
  \item Do not guess—if uncertain, use \texttt{no\_match}.
  \item Always select \textbf{one and only one} category (or \texttt{no\_match}).
\end{itemize}

\textbf{Required Output Format}  
Your response must strictly follow this format:

\begin{verbatim}
CATEGORY: [category number OR "no_match"]
REASON: [1–2 sentences explaining why this category is the best fit]
\end{verbatim}

  \end{tcolorbox}
  \caption{The system prompt for expert classification using a predefined taxonomy.}
  \label{fig:prompt_classification}
\end{figure*}

\begin{figure*}[ht]
  \centering
  \begin{tcolorbox}[
      colback=gray!5!white,
      colframe=dartgreen,
      title=Clustering Inference System Prompt,
      boxrule=0.3mm,
      arc=3mm,
      width=2.0\columnwidth
    ]
\small
You are a helpful clustering assistant. You final answer after reasoning and thinking must follow this exact format for all responses:

\textbf{Required Structure:}

\begin{verbatim}
<answer>
Total clusters: [N]
cluster1: [item1,item2,item3]
cluster2: [item4,item5]
cluster3: [item6,item7,item8]
</answer>
\end{verbatim}

\textbf{Format Rules:}
\begin{enumerate}
  \item Use exactly one answer section
  \item In the answer section, start with "Total clusters: [N]" where N is the number of clusters
  \item List each cluster as "cluster1:", "cluster2:", etc. with items in square brackets
  \item Use 1-based indexing for items (1, 2, 3, ...)
  \item Every item must be assigned to exactly one cluster
  \item No item should appear in multiple clusters or be missing
\end{enumerate}

  \end{tcolorbox}
  \caption{The system prompt for clustering assistant with required output format.}
  \label{fig: prompt_llm_infer_clustering}
\end{figure*}

% \subsection{Prompts used for inference}

\section{Details of Reasoning Chain Generation}
\label{appen:reasoning_chain_generation_detail}

We employ GPT-OSS-20B\footnote{https://openai.com/index/introducing-gpt-oss/} and Phi-4-reasoning\footnote{https://huggingface.co/microsoft/Phi-4-reasoning} agents to generate and verify high-quality reasoning chains: a Generation Agent that produces an initial reasoning trace, and a Review Agent that evaluates and filters it. This setup ensures both the diversity and logical integrity of the final reasoning data. The Generation Agent processes a clustering instruction and a corpus of texts. It outputs a reasoning trace and the final cluster assignments. We observe that without guidance, the models struggle to converge on the correct final cluster assignments. Therefore, to anchor the generation process, we provide the ground-truth answer in the prompt. However, providing the answer introduces the risk of the model "cheating": bypassing genuine reasoning and simply justifying the given solution. To mitigate this, the prompt explicitly instructs the Generation Agent to simulate a first-principles thought process (Figure~\ref{fig: reason_chain_gen}). Despite these constraints, we find that the Generation Agent could still produce flawed or superficial reasoning. Some traces exhibited logical gaps, inconsistent assignments, or simply jumping to the final answer without a credible exploratory and thinking process. To address this, we introduced a Review Agent to serve as an expert evaluator (Figure~\ref{fig: reason_chain_refine}). For each trace, the Review Agent performs an automated assessment across three criteria: (1) completeness of the logical flow, (2) appropriateness of reasoning length, and (3) coherence of stepwise inference. Traces that fail these quality checks are filtered out, ensuring that only high-quality, logically sound reasoning chains are included in our final distillation dataset.

% This automated evaluation greatly improves annotation efficiency and consistency. Empirically, the Review Agent demonstrates strong alignment with human judgments on a held-out subset (agreement rate above 0.91). After filtering, we retain approximately 12% of the original dataset (around 9K examples) as the final distillation set.

\begin{figure*}[ht]
  \centering
  \begin{tcolorbox}[
      colback=gray!5!white,
      colframe=dartgreen,
      title=Reasoning Chain generation Prompt,
      boxrule=0.3mm,
      arc=3mm,
      width=2.0\columnwidth
    ]
\small
\textbf{Tasks} \\
You are a clustering assistant to do text clustering. \\
Given a clustering goal and a list of indexed corpus (|1|. text, |2|. text, |3|. text...), cluster them following the think process below. \\
The clustering results should be in square brackets and use 1-based indexing for both total cluster and items (1, 2, 3, ...).

\vspace{1ex}

\textbf{Think process of the task} \\
First, read through all texts and think how can they be clustered based on the goal, determine the total number of clusters and each cluster's central meaning. \\
Then think about how to assign all texts into these clusters, ensure each text match the cluster's central meaning. \\
Finally, assign each text to a cluster based on the reasoning above. \\
Check the answer format before giving the final answer: every item must be assigned to exactly one cluster, and no item should appear in multiple clusters or be missing.

\vspace{1ex}

\textbf{Goal:} \{instruction\} \\
\textbf{Text:} \\
\{enumerated\_text\}

\vspace{1ex}

The correct final clustering results are: \\
\texttt{<answer>} \\
\{correct\_answer\} \\
\texttt{</answer>}

\vspace{1ex}

Now for the above tasks, please focus on producing a high-quality think process following the above guidance and naturally leads to the correct clusters. \\
\textbf{IMPORTANT!!!}
\begin{enumerate}
  \item Simulate the think process guidance above as if you are clustering for the very first time, without knowing the final answer.
  \item Avoid any phrasing that implies prior knowledge of the final answer (e.g., "looking at the correct answer," "we must match," "the real clusters are").
  \item Do not perform sanity checks against the correct answer. The reasoning should unfold independently.
  \item The correct results should only emerge after exploration, backtracking, and self-verification without over thinking.
  \item Only return the high-quality think process start with ``Think process:''
\end{enumerate}

  \end{tcolorbox}
  \caption{The system prompt for reasoning chain generation.}
  \label{fig: reason_chain_gen}
\end{figure*}

\begin{figure*}[ht]
  \centering
  \begin{tcolorbox}[
      colback=gray!5!white,
      colframe=dartgreen,
      title=Reasoning Chain Evaluation Prompt,
      boxrule=0.3mm,
      arc=3mm,
      width=2.0\columnwidth
    ]
\small
You are an expert AI judge tasked with evaluating the quality and completeness of reasoning chains. 

Please evaluate the following reasoning chain based on these criteria:

\begin{enumerate}
  \item \textbf{Completeness}: Does the reasoning chain generally provide a complete logical flow?
  \item \textbf{Length Appropriateness}: Is the reasoning chain of appropriate length (not unnaturally short or abruptly cut off)?
  \item \textbf{Logical Coherence}: Does the reasoning follow a logical sequence without gaps or jumps?
\end{enumerate}

\textbf{Reasoning Process of Clustering Task:} \\
First, read through all texts and think how can they be clustered based on the goal, determine the total number of clusters and each cluster's central meaning. \\
Then think about how to assign all texts into these clusters, ensure each text match the cluster's central meaning. \\
Finally, assign each text to a cluster based on the reasoning above.

\vspace{1ex}

\textbf{Reasoning Chain to Evaluate:} \\
\{reasoning\_chain\}

\vspace{1ex}

\textbf{Response Format:} \\
Provide your assessment in the following format:

\texttt{ASSESSMENT: [ACCEPT/REJECT]} \\
\texttt{REASON: [Brief explanation of why you accepted or rejected this reasoning chain]}

\vspace{1ex}

\textbf{Examples:}
\begin{itemize}
  \item \textbf{ACCEPT} if the reasoning is complete, logical, appropriately detailed, and reaches a proper conclusion
  \item \textbf{REJECT} if the reasoning is too short, incomplete, illogical, abruptly cut off, or doesn't properly address the question
\end{itemize}

Your assessment:
  \end{tcolorbox}
  \caption{The system prompt for reasoning chain refinement. }
  \label{fig: reason_chain_refine}
\end{figure*}

%\input{tables/tab_clustering_instructions}

% \begin{table}[h]
%     \centering
%     \scalebox{0.9}{
%     \begin{tabular}{c|cccc}
%     \specialrule{0.1em}{0.2em}{0.2em}
%         Task & Name & \#clusters & \#data(small) & \#data(large) \\
%         \hline
%         \multirow{2}{*}{Intent} & Bank77 & 77 & 3,080 & 10,003 \\
%         & CLINC(I) & 150 & 4,500 & 15,000 \\
%         \hline
%         \multirow{2}{*}{Type} & FewRel & 64 & 4,480 & 40,320 \\
%         & FewEvent & 34 & 4,742 & 18,969 \\
%         \hline
%         \multirow{3}{*}{Topic} & StackEx & 121 & 4,156 & 50,000 \\
%         & ArxivS2S & 93 & 3,674 & 50,000 \\
%         & Reddit & 50 & 3,217 & 50,000 \\
%         \hline
%         \multirow{1}{*}{Emotion} & GoEmo & 27 & 5,940 & 23,485 \\
%         \hline
%         \multirow{2}{*}{Domain} & MTOP(D) & 11 & 4,386 & 15,667 \\
%         & Massive(D) & 18 & 2,974 & 11,514 \\
%     \specialrule{0.1em}{0.2em}{0.2em}
%     \end{tabular}}
%     \vspace{-2mm}
%     \caption{Dataset statistics.}
%     \label{tab:dataset}
%     \vspace{-0.5cm}
% \end{table}
\section{Format Constraints for Instruction-Following Clustering}
\label{appen:format_constraints}

This appendix provides a detailed explanation of the format constraints implemented in our instruction-following clustering framework. These constraints ensure that LRMs generate well-structured, parseable, and complete clustering solutions that adhere to our evaluation protocol.

\paragraph{Response Structure Requirements}
The output response should have exactly one occurrence each of \texttt{<think>}, \texttt{</think>}, \texttt{<answer>}, and \texttt{</answer>} tags. The correct tag ordering: $\texttt{<think>} \texttt{</think>} \texttt{<answer>} \texttt{</answer>}$.
No malformed, nested, or missing structural elements.

% \begin{lstlisting}[language=XML,basicstyle=\small\ttfamily]
% <think>
% [model reasoning process]
% </think>
% <answer>
% [clustering solution]
% </answer>
% \end{lstlisting}

\paragraph{Clustering Solution Format}
Within the \texttt{<answer>} section, clustering solutions must adhere to the following canonical format:
\begin{lstlisting}[basicstyle=\small\ttfamily]
Total clusters: [K]
cluster1: [i1, i2, i3, ...]
cluster2: [j1, j2, j3, ...]
...
clusterK: [m1, m2, m3, ...]
\end{lstlisting}
In the answer section, the response needs to start with "Total clusters: [N]" where N is the number of clusters. Then list each cluster as "cluster1:", "cluster2:", etc., with items in square brackets. Use 1-based indexing for items (1, 2, 3, ...).

\paragraph{Completeness and Consistency Validation}
\circled{1} The \texttt{Total clusters} line must be parseable.
\circled{2} Cluster definitions must match the declared count.
\circled{3} All item indices from \texttt{1} to \texttt{N} (total items) must appear exactly once.
\circled{4} Duplicate, missing, or out-of-bound indices result in format errors.
Violations of any of these rules are classified as format errors.

\paragraph{Cluster Count Consistency}
The declared cluster count $K$ must match the actual number of clusters provided for the cluster. This prevents inconsistencies where models declare ``Total clusters: [3]'' but provide only two cluster assignments, ensuring alignment between high-level cluster count and actual cluster assignments.

\section{Baselines Implementation Details}
\label{appen: embedding_methods}

\subsection{Embedding-based Clustering Baselines}
We evaluate several embedding-based clustering methods, following the official implementation for each baseline. For all models, a task-specific prompt is provided to guide the generation of embeddings tailored to the clustering objective. We detail the prompting strategies and implementation specifics below.

\paragraph{Instructor}
We follow the official implementation and parameterization. Instructor is an instruction-tuned embedding model: it takes as input a pair \([\text{instruction}, \text{text}]\). During embedding, the instruction is prepended to guide the semantic intent, e.g., ``\texttt{Represent this text for clustering analysis: \{clustering\_instruction\}}''.
% "Represent this text for clustering analysis: \{user\_instruction\}". 
For example, given the user instruction \emph{``Cluster by research domain''}, the full input to the model for a given document is the pair: [``\texttt{Represent this text for clustering analysis: Cluster by research domain}'', ``\texttt{The abstract discusses AI ethics...}'']. Embeddings are obtained via mean pooling over the final hidden states and are subsequently normalized. 

\paragraph{InBedder}
We adopt the official configuration and implementation from InBedder~\citep{answer-need}. Following the default parameters, we use the recommended \texttt{CausalLMEncoder} backbone, \texttt{BrandonZYW/llama-2-7b-InBedder}. Generation hyperparameters are set to \texttt{temperature=0.6}, \texttt{top\_p=0.9}, and \texttt{max\_new\_tokens=3}. Inputs are formatted using the official template: ``\texttt{Input:\{input\} Instruction:\{instruction\} Response:}'',  
where the placeholders are replaced with the raw text and clustering instruction. Embeddings are taken from the final hidden layer output and normalized prior to clustering.

\paragraph{Qwen Embedding}
Inputs for the Qwen embedding model follow the template ``\texttt{Represent this text for clustering analysis. Task:\{user\_instruction\}  Query:\{text\}}'',  
which symmetrically treats all texts as cluster candidates. Embeddings are extracted from the final hidden state corresponding to the end-of-sequence (EOS) token and normalized. The model supports input lengths of up to $8192$ tokens.

\paragraph{OpenAI Embedding.}
We use OpenAI’s \texttt{text-embedding-3-large} model to generate embeddings via the API. The input prompt concatenates the task description and text:  
``\texttt{Represent this text for clustering. Task: \{user\_instruction\}  Text: \{text\}}''.  
For example: ``\texttt{Represent this text for clustering. Task: Cluster by topic.  Text: The news article covers politics...}''. We use the largest embedding dimension of $3072$.

\paragraph{Sentence Transformers.}
For sentence-level baselines such as \texttt{all-MiniLM-L6-v2}, we prepend a simple task prefix:  
``\texttt{Clustering task: \{user\_instruction\}. Text: \{text\}}''. Embeddings are obtained via mean pooling and normalization.

\paragraph{Clustering and Pipeline.}
Clustering is performed using \texttt{scikit-learn} implementations of \texttt{KMeans} or \texttt{GaussianMixture}, with \texttt{random\_state=43} and the ground-truth cluster count. Inputs exceeding $28\text{k}$ tokens are filtered using \texttt{tiktoken}.

\subsection{LLMs baselines}
To ensure a fair comparison between proprietary models and our proposed approaches, we provide a detailed specification of the required output format. This specification explicitly defines all formatting constraints, thereby guaranteeing that clustering results from both LRMs and LLMs baselines can be consistently and correctly parsed. In addition, we carefully designed and refined the answer-parsing functions to ensure robustness and accuracy during evaluation. The system prompt used for inference is illustrated in Figure~\ref{fig: prompt_llm_infer_clustering}.

\section{Training Config Details}
\label{appendix:implementation_details}

Our training framework is based on \texttt{verl}~\cite{sheng2024verl} and \texttt{trl}~\cite{vonwerra2022trl}. 

\paragraph{Distillation Stage.} We conduct SFT over 510 distilled training examples. Each example is constructed by sampling 30 different text combination from the 17 held-in tasks (see Section~\ref{subsec: c0c1c2 construction}) and then conduct distillation.  We use a fixed batch size of 128 and a micro-batch size of 1, training for a single epoch. To optimize GPU memory usage, we enable gradient checkpointing, FlashAttention, and Adam offloading. The learning rates are set based on the model size: $2\mathrm{e}{-5},\ 1\mathrm{e}{-5},$ and $3\mathrm{e}{-6}$ for models of size $3\text{B},\ 7\text{B},$ and $14\text{B}$, respectively.

\paragraph{RL Stage.}
We perform GRPO optimization over 8.5k training examples, where each example is constructed by sampling text from the 17 held-in tasks (see Section~\ref{subsec: c0c1c2 construction}). 
Training is conducted with a global batch size of 32 and a mini-batch size of 128. We adopt Fully Sharded Data Parallel (FSDP) training. Rollouts are generated using \texttt{vLLM} with a tensor parallelism degree of 4, and GPU memory utilization capped at 50\%. 
Sampling follows the default decoding configuration (temperature = 1.0, top-$p$ = 1.0). KL regularization is applied with a coefficient of $1\times10^{-3}$ and a clipping ratio of 0.2. For each prompt, we sample 5 candidate rollout responses. The maximum input and output sequence lengths are set to 28{,}000 and 4{,}000 tokens, respectively. Learning rates are tuned by model scale: $1\times10^{-6}$ for the 3B and 7B models, and $5\times10^{-7}$ for the 14B model. 
We train the 3B, 7B, and 14B variants on 1, 1, and 2 nodes, respectively, with each node equipped with 8 GPUs.

\section{V-measure Metric}
\label{appen: vmeasure}

The V-measure \citep{vmeasure} is an external clustering evaluation metric based on conditional entropy. It quantifies the agreement between a predicted clustering $\hat{C} = \{\hat{C}_1, \dots, \hat{C}_k\}$ and a gold standard partition $C^\ast = \{C^\ast_1, \dots, C^\ast_m\}$ in terms of two complementary criteria:
\begin{itemize}[leftmargin=*]
    \item \textbf{Homogeneity} ($h$): each predicted cluster should contain only members of a single gold class.
    \item \textbf{Completeness} ($c$): all members of a given gold class should be assigned to the same predicted cluster.
\end{itemize}
Formally, let $H(C^\ast)$ denote the entropy of the gold labels and $H(C^\ast\mid\hat{C})$ the conditional entropy given the predicted clustering. 
Homogeneity and completeness are defined as:

\begin{equation}
h =
\begin{cases}
1, & \text{if } H(C^\ast) = 0, \\[4pt]
1 - \dfrac{H(C^\ast \mid \hat{C})}{H(C^\ast)}, & \text{otherwise.}
\end{cases}
\end{equation}

\begin{equation}
c =
\begin{cases}
1, & \text{if } H(\hat{C}) = 0, \\[4pt]
1 - \dfrac{H(\hat{C} \mid C^\ast)}{H(\hat{C})}, & \text{otherwise.}
\end{cases}
\end{equation}
The V-measure is then given by the harmonic mean:
\begin{equation}
V = \frac{2 \cdot h \cdot c}{h + c}.
\end{equation}
This symmetric formulation ensures that a high $V$ is achieved only when clusters are both internally consistent (high $h$) and well aligned with the gold classes (high $c$). Values range from $0$ (no agreement) to $1$ (perfect match).

\section{Details of Data Pre-processing}
\label{appendix: data preprocess}

\subsection{LMSYS-Chat}
\label{appendix: LMSYS-Chat}
Current intent-discovery benchmarks such as Bank77 and CLINC~\citep{li-etal-2021-mtopintent,Casanueva2020bank77intent,larson-etal-2019-CLINICintent,fitzgerald2022massiveintent} primarily consist of short, single-turn utterances associated with narrowly defined intents. While useful for intent classification, these datasets fail to capture the complexity of real-world interactions, where dialogues are longer, context-dependent, and often involve nuanced shifts in user goals. To address this gap, we leverage the LMSYS-Chat-1M corpus~\citep{zheng2023lmsys}, which contains over one million real-world conversations collected from more than 210,000 unique users through the Vicuna demo and Chatbot Arena~\citep{zheng2023judgingvicuna}. For our benchmark, we focus on the first 500,000 English-language dialogues and apply rigorous filtering to exclude unsafe or low-quality content. Although multi-turn conversations better reflect natural dialogue, they frequently exhibit shifting or overlapping intents that complicate cluster assignment. Therefore, we restrict our selection to one- and two-turn exchanges, preserving conversational richness while ensuring clearer intent detection. Distinct from prior intent-clustering datasets that focus solely on user utterances, we also incorporate the assistant's responses into the clustering process. For example, we introduce clustering dimensions based on expected output formats, thereby capturing a broader range of realistic use cases beyond user intent alone. This dual perspective acknowledges that effective clustering in human–AI dialogue requires jointly modeling user communicative goals and the structural or stylistic properties of system replies. Given the scale of the corpus, we implement our multi-agent pipeline hierarchically. In the first stage, the pipeline groups dialogues into 12 coarse-grained intent clusters. In the second stage, each cluster is further subdivided into multiple sub-clusters, yielding a more granular taxonomy of conversational patterns. This hierarchical strategy balances scalability with interpretability and supports downstream applications in dialogue understanding and controllable generation. An illustrative example is shown in Figure~\ref{fig:lmsys_example}.

\begin{figure*}[ht]
  \centering
  \begin{tcolorbox}[
      colback=gray!5!white,
      colframe=dartgreen,
      title=LMSYS-Chat Data Example,
      boxrule=0.3mm,
      arc=3mm,
      width=2.1\columnwidth
    ]
\small

\textbf{User:}  
What features are relevant in a visualization of a streaming platform? (From a business intelligence viewpoint)

\vspace{1ex}
\textbf{Assistant:}  
When it comes to visualizing a streaming platform from a business intelligence viewpoint, several key features are relevant:  

1. \emph{User engagement} — number of active users, average time spent, number of streams viewed.  
2. \emph{Content distribution} — number of channels/streams, countries/regions available, devices/platforms supported.  
3. \emph{Revenue and monetization} — number of paying subscribers, ARPU, ad views/clicks.  
4. \emph{User acquisition and retention} — cost of acquiring a user, number of new users, retention rates.  
5. \emph{Technical performance} — uptime, load times, outages/issues. These visualizations provide insights into platform performance and growth, enabling informed decision-making.  

\vspace{2ex}
\textbf{User:}  
Relevant features when comparing content streaming platforms, such as Netflix vs Disney+

\vspace{1ex}
\textbf{Assistant:}  
When comparing platforms like Netflix and Disney+, the following features matter:  

1. \emph{User engagement} — active users, time spent, streams viewed.  
2. \emph{Content library} — total titles, originals, exclusives.  
3. \emph{Geographic coverage} — countries/regions available, supported devices/platforms.  
4. \emph{Pricing and monetization} — pricing strategy, subscribers, ARPU, ad revenues.  
5. \emph{Marketing and promotion} — budget, partnerships, user acquisition.  
6. \emph{Technical performance} — uptime, stream load time, outages.  

These comparisons yield insights into competitiveness and positioning in the market.  

  \end{tcolorbox}
  \caption{LMSYS-Chat data example.}
  \label{fig:lmsys_example}
\end{figure*}

\subsection{ECHR}
\label{appendix: ECHR}

We incorporate case law documents from the European Court of Human Rights (ECHR)~\citep{chalkidis-etal-2019-echr}, which introduce long-form, argument-rich texts that pose distinct reasoning challenges compared to conversational data. Each case typically comprises: (i) a facts section organized into numbered paragraphs summarizing events and evidence; (ii) the alleged violations of the European Convention on Human Rights; and (iii) the court’s decision indicating which allegations are upheld or dismissed. We use an enriched version of the publicly available ECHR corpus containing approximately 11,000 cases. To ensure data quality and consistency, we exclude non-English cases and those lacking sufficient information across the core components (facts, articles, or decisions). To facilitate effective clustering based on legal reasoning rather than superficial features, we preprocess each case from the ECHR corpus by removing metadata fields that are irrelevant to the core legal content, such as "Importance Level" and "Judges." We also omit the conclusion and the explicitly stated violated articles. This ensures that models must engage in substantive understanding and normative reasoning to infer which articles of the European Convention on Human Rights may have been violated. Clustering is then performed using the instruction: \emph{“Cluster legal cases by the violated article.”} An illustrative example is shown in Figure~\ref{fig:schvarc_case}, which presents a structured summary of the case \emph{Schvarc v. Slovakia} (2014) prior to redaction. The “Facts” section outlines the legal proceedings, while conclusion and violated articles are removed during training to simulate realistic inference conditions.

\begin{figure*}[ht]
  \centering
  \begin{tcolorbox}[
      colback=gray!5!white,
      colframe=dartgreen,
      title=ECHR Data Example,
      boxrule=0.3mm,
      arc=3mm,
      width=2.1\columnwidth
    ]
\small

\textbf{Case Name:} CASE OF SCHVARC v. SLOVAKIA  
\textbf{Item ID:} 001-139999  
\textbf{Respondent:} Slovakia (SVK)  
\textbf{Court Branch:} Chamber  
\textbf{Date:} 2014  
\textbf{Judges:} Alvina Gyulumyan; Ján Šikuta; Johannes Silvis; Josep Casadevall; Kristina Pardalos; Luis López Guerra  
\textbf{Importance Level:} 4  

\vspace{1ex}
\textbf{Facts:}  
The applicant, born in 1978, resides in Krupina. Arrested, charged, and remanded in custody on 20–22 May 2009 for drug-related offences under Article 172 §1(c) of the Criminal Code. On 22 July 2009, he requested release, pledging lawful conduct and non-interference with justice. On 3 August 2009, the District Prosecutor recommended dismissal; the Zvolen District Court dismissed the request on 19 August 2009, citing a prior theft conviction. The applicant appealed orally on 19 August, later amending with legal assistance. Written decision served on 24 September 2009. On 8 October 2009, the Banská Bystrica Regional Court dismissed the appeal; written version served on 30 October 2009. On 11 November 2009, the applicant lodged a complaint with the Constitutional Court under Article 127 of the Constitution and Article 5 §4 of the Convention, claiming €1,500. The Constitutional Court admitted the complaint on 18 February 2010 and ruled on 30 March 2010, finding a violation of the right to speedy review and awarding €500 compensation. Written judgment served on 24 May 2010.  

\vspace{1ex}
\textbf{Conclusion:}  
- \emph{Violation of Article 5} – Right to liberty and security (Article 5-4: Speediness of review).  

\vspace{1ex}
\textbf{Violated Articles/Paragraphs:}  
- Article 5 (para. 4)  

  \end{tcolorbox}
  \caption{ECHR data example.}
  \label{fig:schvarc_case}
\end{figure*}

\subsection{S\&P~500 Financial Reports}
\label{appendix: sp500}
To broaden coverage beyond conversational and legal texts, we include financial documents in the form of public company annual reports (Form 10-K filings). Specifically, we leverage the SP500-EDGAR-10K dataset\footnote{\url{https://huggingface.co/datasets/jlohding/sp500-edgar-10k}}, which contains reports for all historical S\&P~500 constituents from 2010 to 2022~\cite{sec_edgar}. Each report consists of multiple structured items (e.g., Business, Risk Factors, Management’s Discussion \& Analysis, Market Risk Disclosures, Financial Statements). To ensure compatibility with context-length constraints during model training and inference, we select a subset of SP500 financial filings in which the length of each text item remains within a tractable range. This allows multiple document segments to be processed jointly without exceeding the model’s input limits. We focus on three representative items from the 10-K filings:
\begin{itemize}[leftmargin=*]
    \item \textbf{Item~2 – Properties:} Describes corporate assets such as manufacturing plants, distribution centers, and office facilities, including their location, usage, and ownership status.
    \item \textbf{Item~3 – Legal Proceedings:} Summarizes ongoing or pending litigation and regulatory investigations. By aligning structurally with ECHR case law—though differing substantially in domain—this item supports evaluation of cross-domain generalization in legal clustering tasks.
    \item \textbf{Item~5 – Market for Registrant’s Common Equity:} Covers stock exchange listings, price histories, dividends, and shareholder matters. Including this item enables assessment of model robustness when transferred to financial topics unrelated to legal reasoning.
\end{itemize}
An example of Item~2 (“Properties”) is presented in Figure~\ref{fig:company_properties_2010}.

\begin{figure*}[ht]
  \centering
  \begin{tcolorbox}[
      colback=gray!5!white,
      colframe=dartgreen,
      title=SP500 Data Example,
      boxrule=0.3mm,
      arc=3mm,
      width=2.0\columnwidth
    ]
\small

\textbf{Item:} 2 – Properties  

\vspace{1ex}
\textbf{General Overview:}  
- Company and subsidiaries owned or leased material facilities (over 50,000 sq. ft.) in 16 U.S. states and 14 foreign countries.  
- Facilities used for manufacturing, distribution, and sales offices.  
- Properties used by multiple segments are attributed to the segment with highest usage.  
- Facilities considered suitable and adequate for business.  

\vspace{1ex}
\textbf{Owned Material Facilities:}  
- \emph{Security:} Farmington (CT); Sterling, Rock Falls (IL); Indianapolis (IN); Nicholasville (KY); Richmond (VA); Cobourg (Canada); Nueva Leon (Mexico); Xiaolan (China).  
- \emph{Industrial:} Phoenix (AZ); Two Harbors (MN); Columbus, Georgetown, Sabina (OH); Allentown (PA); Dallas (TX); Pecky (Czech Rep.); Epernay, Ezy Sur Eure, Feuquieres en Vimeu, Morangis, Villeneuve Le Roi (France); Fano, Gemonio, Monvalle (Italy).  
- \emph{CDIY:} Clinton, New Britain (CT); Shelbyville (IN); East Greenwich (RI); Cheraw (SC); Smiths Falls (Canada); Hellaby, Northampton (England); Arbois, Besancon Cedex, Lassiey (France); Puebla (Mexico); Jiashan City, Langfang, Xiaolan (China); Wroclaw (Poland); Taichung Hsien (Taiwan); Amphur Bangpakong (Thailand).  

\vspace{1ex}
\textbf{Leased Material Facilities:}  
- \emph{Corporate Offices:} New Britain (CT).  
- \emph{Security:} Noblesville (IN); Kentwood (MI).  
- \emph{Industrial:} Highland Heights, Westerville (OH); Milwaukie (OR); Morangis (France).  
- \emph{CDIY:} Miramar (FL); Fishers (IN); Kannapolis (NC); Epping (Australia); Mechelen (Belgium); Oakville (Canada); Leeds (England); Karmiel, Migdal (Israel); Biassono, Figino Serenza (Italy); Pietermaritzburg (South Africa).  

\vspace{1ex}
\textbf{Facilities Not in Use:}  
- \emph{Security:} Richmond (VA, owned).  
- \emph{Industrial:} Ezy Sur Eure, Villeneuve Le Roi (France, owned).  
- \emph{CDIY:} Clinton (CT, owned); Smiths Falls (Canada, owned); Leeds (England, leased); Amphur Bangpakong (Thailand, one property owned).  

  \end{tcolorbox}
  \caption{Example of Item~2 (“Properties”) from SP500 dataset.}
  \label{fig:company_properties_2010}
\end{figure*}

\subsection{Benchmark Details}
\label{appen: benchmark details}
In Table~\ref{tab: clustering instructions}, we present the clustering instructions for both held-in and held-out tasks across each dataset. We used GPT-OSS-20B\footnote{https://openai.com/index/introducing-gpt-oss/} as the LLM agent for different parts of benchmark generation pipeline, including LLM judges.

\begin{table*}[ht!]
    \centering
    \scalebox{0.80}{
    \begin{tabularx}{1.2\textwidth}{c|c|X}
    \specialrule{0.1em}{0.2em}{0.2em}
        Dataset & Split & Prompt \\
        \hline
        \multirow{2}{*}{ECHR} 
            & Held-in  & 
                Cluster these legal cases by the primary vulnerable population involved. \newline
                Cluster these legal cases by the dominant type of domestic proceedings that preceded the European Court application. \newline
                Cluster these legal cases by the violation articles invoked in the Court's judgment. \\
        \cline{2-3}
            & Held-out & 
                Cluster the following texts based on the primary underlying category of legal dispute or issue. \newline
                Cluster the following texts based on the main types of parties or entities involved as adversaries or plaintiffs. \newline
                Cluster the following texts based on the timing, duration, or current stage of resolution for the described legal proceedings. \\
        \hline
        \multirow{2}{*}{LMSY} 
            & Held-in & 
                Cluster the following user requests based on explanation audience level. \newline
                Cluster these requests by temporal regime they focus on. \newline
                Cluster the dialogue by identifying the root cause of the issue. \newline
                Cluster the dialogue based on the type of transformation needed to convert it into a different form or function. \newline
                Cluster these creative generation requests by content type. \newline
                Group each dialogue according to the primary social function or interpersonal purpose it serves. \newline
                Cluster these decision support requests based on the structure and format of the solution being requested. \newline
                Group these recommendation requests by identifying the type of justification used. \newline
                Cluster the dialogues by the distinct perspective or evaluative role the agent adopts. \newline
                Group the conversations based on the type of structured planning or scheduling being discussed. \\ 
        \cline{2-3}
            & Held-out & 
                Group these computational tasks by archetype. \newline
                Cluster these data analysis requests by tooling context. \newline
                Cluster these texts based on the kind of external context or disambiguation needed. \newline
                Group these factual lookup requests by the structure and format of the expected output. \newline
                Cluster these factual lookup requests based on the type of source or origin of information. \\
        \hline
        \multirow{2}{*}{SP500} 
            & Held-in  & 
                Cluster these SP500 companies by their asset ownership and utilization patterns. \newline
                Cluster these SP500 companies by their geographic distribution of assets and operations. \newline
                Cluster these SP500 companies by their industry-specific asset composition. \newline
                Cluster these SP500 companies by their risk exposure and asset encumbrance patterns. \\ 
        \cline{2-3}
            & Held-out & 
                Cluster the texts based on the strategies companies use for distributing earnings to shareholders via dividends. \newline
                Cluster the texts based on the approaches companies take toward buying back their own shares. \newline
                Cluster the texts based on the varieties of equity instruments issued by companies. \\
    \specialrule{0.1em}{0.2em}{0.2em}
    \end{tabularx}
    }
    \caption{Clustering instructions for both held-in and held-out tasks from each dataset.}
    \label{tab: clustering instructions}
\end{table*}

\section{Analysis of Main Results}
\label{appen: analyze_main_results}

\subsection{Format Errror Analysis}
\begin{figure*}[ht]
    \centering
    % \resizebox{\linewidth}{!}{
    \includegraphics[width=2.0\columnwidth]{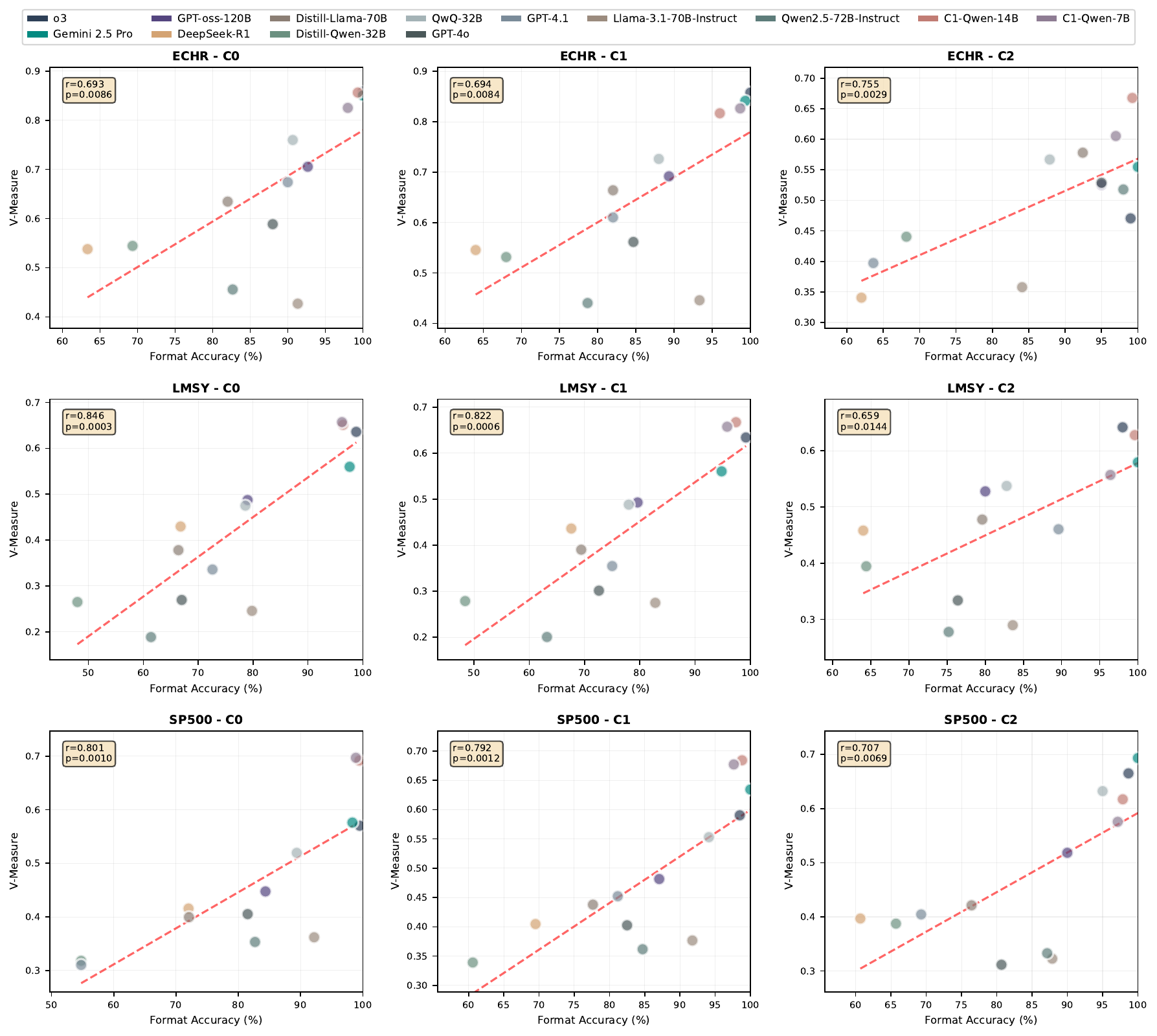}
    % }
    \caption{Relationship between format accuracy and V-measure score. Models with higher format accuracy consistently achieve higher V-measure scores, highlighting format correctness as a fundamental prerequisite for clustering quality. Nevertheless, high format accuracy alone does not guarantee superior clustering performance. For instance, models such as Llama-3.1-70B-Instruct exhibit high format accuracy but still attain low V-measure scores. Effective clustering further depends on a model’s capacity for semantic reasoning and clustering instruction alignment. Mastering the syntax of clustering outputs is necessary but not sufficient.}
    \label{fig: grid}
\end{figure*}

As shown in Figure \ref{fig: grid}, models with higher format accuracy consistently achieve higher V-measure scores, highlighting format correctness as a fundamental prerequisite for clustering quality. This correlation arises because an incorrectly formatted response that cannot be parsed receives a V-measure score of $0$. Consequently, general-purpose language models often underperform, as they frequently violate the required output format. Nevertheless, high format accuracy alone does not guarantee superior clustering performance. For instance, models such as Llama-3.1-70B-Instruct exhibit high format accuracy but still attain low V-measure scores. This discrepancy suggests that while high format accuracy ensures evaluability, effective clustering further depends on a model’s capacity for semantic reasoning and clustering instruction alignment. Mastering the syntax of clustering outputs is necessary but not sufficient. The model must also internalize the underlying reasoning process that drives coherent and instruction-faithful cluster assignment.

\subsection{Case Study}
Figure~\ref{fig:goodcase} illustrates a representative success case of our clustering agent on the ECHR dataset, highlighting the model’s ability to perform interpretable reasoning while maintaining a structural output format. In this example, the model first infers an appropriate number of clusters and articulates a clear rationale for each, corresponding to distinct categories of human rights violations such as Right to Privacy, Protection of Property, Freedom of Expression, and Right to a Fair Trial. The reasoning sequence reveals a deliberate, hierarchical decision process in which the model identifies the underlying legal principle in each text segment before assigning it to a cluster. The resulting clusters are both logically coherent and legally interpretable, demonstrating that the model not only adheres to output format conventions but also internalizes domain-specific reasoning patterns. This example shows reasoning supervision enables models to construct clusters that are semantically meaningful, structurally consistent, and faithful to the task instruction.

% Crucially, the model’s reasoning (<think>) exhibits a structured chain of thought aligned with legal semantics rather than superficial lexical similarity. It differentiates between closely related concepts—for instance, distinguishing Freedom of Assembly from Freedom of Association—and consistently maps textual evidence to the relevant article of the European Convention on Human Rights. 
\subsection{Failure Case}

Figure~\ref{fig:badcase} illustrates a representative failure case where the clustering agent exhibits brittle reasoning and unstable self-revision. The instruction is to group ECHR cases by the primary vulnerable population. The agent initially proposes sensible, mutually exclusive categories and begins assigning texts accordingly. During a later “reflection” pass, however, it merges Children with a broader bucket (Victims of discrimination based on social status) and then continues to use Children as a standalone label in assignments. This revision creates overlapping categories and internal inconsistency: Children is both a subtype inside a broad “Vulnerable individuals” bucket and a separate top-level cluster used for item assignment. One possible reason is our current reinforcement objective optimizes final-format validity and end-state clustering quality, but it does not explicitly reward the process of reasoning. Concretely, the reward is the sum of a format term and a clustering term with count accuracy and V-measure, without intermediate checks on reasoning steps or on constraints like mutual exclusivity and consistent granularity. In this case, the agent’s “reflection” collapses distinct categories, introduces overlap, and then assigns items inconsistently.

% Example box
\onecolumn
\begin{tcolorbox}[
  enhanced jigsaw,
  breakable,
  colback=gray!5!white,
  colframe=dartgreen,
  title={Clustering Agent Response Example (Full Reasoning and Answer)},
  fonttitle=\footnotesize,
  boxrule=0.3mm,
  arc=3mm,
  width=\textwidth,         % full page width (both columns in a starred float)
]
\footnotesize
\textbf{\textcolor{dartgreen}{\uline{First, I will read through all texts to understand the core legal issue in each case, keeping the goal in mind: "Cluster these legal cases by the violation articles invoked in the Court's judgment." I need to identify the specific right or principle of the European Convention on Human Rights that is at stake. Let's break down each text:}}}

\begin{itemize}[itemsep=1pt,] % topsep=0pt
  \item \textbf{Text 1 (JAEGER v. ESTONIA):} The applicant was searched in a public area, and the door was transparent. This is a clear issue of privacy and dignity under Article 8 (Right to respect for private and family life).
  \item \textbf{Text 2 (ALTIN AND OTHERS v. TURKEY):} The applicants were unable to get a judgment enforced. This is a problem with the right to a fair trial and the right to peaceful enjoyment of possessions (the judgment award). It falls under Article 1 of Protocol No. 1 (Protection of property) and Article 13 (Right to an effective remedy).
  \item \textbf{Text 3 (Massimo ROSA v. ITALY):} The applicant was unable to get a court order enforced to evict a tenant. This is another case about the right to peaceful enjoyment of possessions under Article 1 of Protocol No. 1.
  \item \textbf{Text 4 (STREHAR v. SLOVENIA):} The case concerns the length of pre-trial detention. This is a classic Article 5 (Right to liberty) issue.
  \item \textbf{Text 5 (GAWEDA v. POLAND):} The government refused to register a periodical based on its title. This is a restriction on freedom of the press under Article 10 (Freedom of expression).
  \item \textbf{Text 6 (UNSAL OZTURK v. TURKEY):} The applicant's books were confiscated based on their content. This is another case concerning freedom of the press under Article 10.
  \item \textbf{Text 7 (UNION EUROPEENNE DES DROITS DE L'HOMME AND JOSEPHIDES v. TURKEY):} Security forces seized the applicants' flag and confiscated their materials during a peaceful demonstration. This is an interference with the right to peaceful assembly under Article 11.
  \item \textbf{Text 8 (SIVERIN v. RUSSIA):} The applicants' judgment for benefits was not enforced. Like text 2, this is another Article 1 of Protocol No. 1 (Protection of property) and Article 13 issue.
  \item \textbf{Text 9 (KHALILOVA AND AYYUBZADE v. AZERBAIJAN):} The applicants were arrested and prosecuted for participating in a peaceful protest. This is another Article 11 (Freedom of assembly) issue.
  \item \textbf{Text 10 (SZEMELY ES VAGYONOROK FUGGETLEN SZAKSZERVEZETI SZOVETSEGE v. HUNGARY):} A trade union was refused registration. This is another case concerning freedom of association under Article 11.
  \item \textbf{Text 11 (DURMUS v. TURKEY):} A teacher was sanctioned for distributing a union notice. This is another Article 11 issue, concerning the freedom of association.
  \item \textbf{Text 12 (CLOOTH v. BELGIUM):} The applicant's claim for compensation was dismissed. The issue is the right to an effective remedy under Article 13.
  \item \textbf{Text 13 (T.M. AND OTHERS v. RUSSIA):} The applicants face removal to Uzbekistan where they fear ill-treatment. This is an Article 3 (Prohibition of inhuman or degrading treatment) issue.
  \item \textbf{Text 14 (PEREE v. THE NETHERLANDS):} The applicant was prosecuted for statements made in a political interview. This is another Article 10 (Freedom of expression) issue.
  \item \textbf{Text 15 (NIEMINEN v. FINLAND):} The police entered the applicant's home to investigate a crime. This is another Article 8 issue concerning the right to privacy.
  \item \textbf{Text 16 (ALBAYRAK v. TURKEY):} The applicant was transferred to a lower court for trial. The justification was his "undermine[d] the honour and dignity of the judiciary." This concerns the right to a fair trial under Article 2 (Right to a fair trial).
  \item \textbf{Text 17 (DEMIR v. FINLAND):} The applicant was expelled, separating him from his family. This is another Article 8 issue concerning the right to respect for his private life and family life.
  \item \textbf{Text 18 (NIYITEGURE v. SWEDEN):} The applicant fears persecution if deported to Burundi and seeks asylum. This is another Article 3 issue concerning the risk of inhuman or degrading treatment.
\end{itemize}
\textbf{\textcolor{dartgreen}{\uline{Based on this analysis, I can identify several distinct categories of legal issues:}}}
\begin{enumerate}[itemsep=1pt,]
  \item \textbf{Privacy and Dignity:} Cases involving searches or invasions of personal space.
  \item \textbf{Property and Compensation:} Cases where the state fails to protect property or enforce judgments.
  \item \textbf{Freedom of Expression:} Cases concerning restrictions on the press or speech.
  \item \textbf{Freedom of Assembly:} Cases concerning the right to protest.
  \item \textbf{Freedom of Association:} Cases concerning the right to form or join organizations.
  \item \textbf{Right to a Fair Trial:} Cases concerning procedural fairness in criminal proceedings.
  \item \textbf{Right to Prohibition of Inhuman or Degrading Treatment:} Cases concerning the risk of ill-treatment.
\end{enumerate}

This suggests a total of 7 clusters. I will define them as follows:
\begin{itemize}
  \item \textbf{Cluster 1: Right to Privacy and Dignity (Article 8).} Cases involving searches or other state intrusions into personal space.
  \item \textbf{Cluster 2: Protection of Property and Effective Remedy (Article 1 of P1-1 \& 13).} Cases about the state's failure to protect assets or provide a proper remedy.
  \item \textbf{Cluster 3: Freedom of Expression (Article 10).} Cases concerning restrictions on media or speech.
  \item \textbf{Cluster 4: Freedom of Assembly (Article 11).} Cases concerning the right to protest.
  \item \textbf{Cluster 5: Freedom of Association (Article 11).} Cases concerning the right to join organizations.
  \item \textbf{Cluster 6: Right to a Fair Trial (Article 2).} Cases concerning the fairness of legal proceedings.
  \item \textbf{Cluster 7: Prohibition of Inhuman or Degrading Treatment (Article 3).} Cases concerning the risk of ill-treatment.
\end{itemize}
\textbf{\textcolor{dartgreen}{\uline{Now, I will go through each text and assign it to one of these clusters based on my reasoning above.}}}
\begin{itemize}[itemsep=1pt,]
  \item Text 1 $\rightarrow$ Issue is privacy/dignity $\rightarrow$ \textbf{Cluster 1}.
  \item Text 2 $\rightarrow$ Issue is property/compensation enforcement $\rightarrow$ \textbf{Cluster 2}.
  \item Text 3 $\rightarrow$ Issue is property/compensation enforcement $\rightarrow$ \textbf{Cluster 2}.
  \item Text 4 $\rightarrow$ Issue is liberty/pre-trial detention $\rightarrow$ \textbf{Cluster 6}.
  \item Text 5 $\rightarrow$ Issue is freedom of expression (press) $\rightarrow$ \textbf{Cluster 3}.
  \item Text 6 $\rightarrow$ Issue is freedom of expression (press) $\rightarrow$ \textbf{Cluster 3}.
  \item Text 7 $\rightarrow$ Issue is freedom of assembly $\rightarrow$ \textbf{Cluster 4}.
  \item Text 8 $\rightarrow$ Issue is property/compensation enforcement $\rightarrow$ \textbf{Cluster 2}.
  \item Text 9 $\rightarrow$ Issue is freedom of assembly $\rightarrow$ \textbf{Cluster 4}.
  \item Text 10 $\rightarrow$ Issue is freedom of association $\rightarrow$ \textbf{Cluster 5}.
  \item Text 11 $\rightarrow$ Issue is freedom of association $\rightarrow$ \textbf{Cluster 5}.
  \item Text 12 $\rightarrow$ Issue is lack of an effective remedy $\rightarrow$ \textbf{Cluster 2}.
  \item Text 13 $\rightarrow$ Issue is prohibition of inhuman/degrading treatment $\rightarrow$ \textbf{Cluster 7}.
  \item Text 14 $\rightarrow$ Issue is freedom of expression (speech) $\rightarrow$ \textbf{Cluster 3}.
  \item Text 15 $\rightarrow$ Issue is privacy/home search $\rightarrow$ \textbf{Cluster 1}.
  \item Text 16 $\rightarrow$ Issue is fairness of trial (judicial integrity) $\rightarrow$ \textbf{Cluster 6}.
  \item Text 17 $\rightarrow$ Issue is prohibition of inhuman/degrading treatment (expulsion) $\rightarrow$ \textbf{Cluster 7}.
  \item Text 18 $\rightarrow$ Issue is prohibition of inhuman/degrading treatment (deportation) $\rightarrow$ \textbf{Cluster 7}.
\end{itemize}

The assignments are complete. I will now format the final answer.

% \vspace{1ex}
\texttt{<answer>} \\
\texttt{Total clusters: [7]} \\
\texttt{cluster1: [1, 15]} \\
\texttt{cluster2: [2, 3, 8, 12]} \\
\texttt{cluster3: [5, 6, 14]} \\
\texttt{cluster4: [7, 9]} \\
\texttt{cluster5: [10, 11]} \\
\texttt{cluster6: [4, 16]} \\
\texttt{cluster7: [13, 17, 18]} \\
\texttt{</answer>}

\end{tcolorbox}
\captionof{figure}{An illustrative example of the clustering agent's full reasoning and final structured output for ECHR legal cases clustered by violation articles.}
\label{fig:goodcase}
%   \caption{An illustrative example of the clustering agent's full reasoning and final structured output for ECHR 
%   legal cases clustered by violation articles.}
%   

\begin{tcolorbox}[
enhanced jigsaw,
breakable,
colback=gray!5!white,
colframe=red!70!black,
title={Clustering Agent Failure Case},
fonttitle=\footnotesize,
boxrule=0.3mm,
arc=3mm,
width=\textwidth,
]
\footnotesize
First, I will read through all texts to understand the nature of the cases and identify potential patterns or themes. The goal is to cluster them by the \textbf{"primary vulnerable population involved."}
This means I need to identify the specific demographic or social group that faces heightened human rights risks in each case. After reading through the texts, I can identify several potential clusters:
\begin{itemize}[itemsep=1pt]
\item \textbf{Children:} Cases where the vulnerable population is a child.
\item \textbf{Journalists:} Cases where the vulnerable population is a journalist.
\item \textbf{People with disabilities:} Cases where the vulnerable population is a person with a disability.
\item \textbf{Elderly people:} Cases where the vulnerable population is an elderly person.
\item \textbf{Victims of domestic violence or sexual abuse:} Cases where the vulnerable population is a victim of domestic violence or sexual abuse.
\item \textbf{Victims of discrimination based on social status:} Cases where the vulnerable population is a person facing discrimination based on social or economic status (e.g., victims of domestic conflict, low-income individuals).
\end{itemize}

Next, I will go through each text and assign it to one of these clusters.

\begin{itemize}[itemsep=1pt]
\item \textbf{Text 1:} The applicant is a 19-year-old woman seeking social welfare assistance.
→ This case involves a young person (a child) who is vulnerable due to her age and economic situation.
\item \textbf{Text 2:} The applicant is a journalist convicted for publishing a newspaper.
→ This case involves a journalist.
\item \textbf{...}
\end{itemize}
Based on this analysis, I can define the following clusters:
% \vspace{1ex}
% \textbf{\textcolor{red!70!black}{\uline{Based on this analysis, I can define the following clusters:}}}

\begin{enumerate}[itemsep=1pt]
\item \textbf{Children:} Cases where the vulnerable population is a child.
\item \textbf{Journalists:} Cases where the vulnerable population is a journalist.
\item \textbf{People with disabilities:} Cases where the vulnerable population is a person with a disability.
\item \textbf{Elderly people:} Cases where the vulnerable population is an elderly person.
\item \textbf{Victims of domestic violence or sexual abuse:} Cases where the vulnerable population is a victim of domestic violence or sexual abuse.
\item \textbf{Victims of discrimination based on social status:} Cases where the vulnerable population is a person facing discrimination based on social status (e.g., victims of domestic conflict or poverty).
\end{enumerate}

% \vspace{1ex}
% \textbf{\textcolor{red!70!black}{\uline{Revision and Error Recognition:}}}

\textbf{\textcolor{red!70!black}{\uline{However, after re-evaluating the texts, I realize that text 4 (a woman who receives welfare payments for her child) and text 11 (a woman living with her daughter in indecent and insalubrious conditions) fit better into the cluster of "Victims of discrimination based on social status" rather than "Children." Therefore, I will merge the "Children" cluster and the "Victims of discrimination based on social status" cluster into a single cluster called "Vulnerable individuals (children, elderly, or victims of discrimination based on social status)."}}}

\vspace{1ex}
This gives me five clusters:

\begin{itemize}[itemsep=1pt]
\item \textbf{\textcolor{red!70!black}{\uline{Children:}}} Cases where the vulnerable population is a child.
\item \textbf{Journalists:} Cases where the vulnerable population is a journalist.
\item \textbf{People with disabilities:} Cases where the vulnerable population is a person with a disability.
\item \textbf{Elderly people:} Cases where the vulnerable population is an elderly person.
\item \textbf{Vulnerable individuals (\textbf{\textcolor{red!70!black}{\uline{children}}}, elderly, or victims of discrimination based on social status):}
Cases where the vulnerable population is a child, an elderly person, or someone discriminated against due to social or economic status.
\end{itemize}

% \vspace{1ex}
% \textbf{\textcolor{red!70!black}{\uline{Final Assignment of Texts:}}}

% \begin{itemize}[itemsep=1pt]
% \item \textbf{Text 1:} The applicant is a 19-year-old woman. → Fits into the \textbf{"Children"} cluster.
% \item \textbf{Text 2:} The applicant is a journalist. → Fits into the \textbf{"Journalists"} cluster.
% \item \textbf{...}
% \end{itemize}

\vspace{1ex}
\textit{\textcolor{red}{(Note: During the “reflection” process, the model merges Children with a broader bucket (Victims of discrimination based on social status) and then continues to use Children as a standalone label in assignments.)}}

\end{tcolorbox}

\captionof{figure}{Example of a failure case where the clustering agent exhibits brittle reasoning and unstable self-revision.}
\label{fig:badcase}

\twocolumn

\end{document}